\documentclass[runningheads]{llncs}
\usepackage{graphicx}

\usepackage{tikz}
\usepackage{comment}
\let\proof\relax

\usepackage{amsmath,amssymb} 
\usepackage{color}

\usepackage[accsupp]{axessibility}  

\usepackage{float} 
\usepackage{subfigure} 
\usepackage{smile}
\usepackage{bbm}
\usepackage{color}
\usepackage{wrapfig}
\usepackage{enumitem}

\usepackage{tabularx} 

\usepackage{listings}

\newcommand\s{0.95cm}

\newcommand{\eg}{\textit{e}.\textit{g}.}

\newcommand{\ifcomments}{\iftrue}

\newcommand{\tabincell}[2]{\begin{tabular}{@{}#1@{}}#2\end{tabular}}  
\newcommand{\rulesep}{\unskip\ \vrule\ }

\begin{document}
\pagestyle{headings}
\mainmatter
\def\ECCVSubNumber{1395}  

\title{Towards Efficient Adversarial Training on Vision Transformers}

\titlerunning{Towards Efficient Adversarial Training on Vision Transformers}
%
\author{Boxi Wu\thanks{Equal contribution.}\inst{1} \and
Jindong Gu$^{\star}$\inst{2} \and
Zhifeng Li\inst{3} \and
Deng Cai\inst{1} \and
Xiaofei He\inst{1} \and
Wei Liu\inst{3}
} 
\authorrunning{B. Wu et al.}
%
\institute{State Key Lab of CAD\&CG, Zhejiang University \\
\email{boxiwu@zju.edu.cn},
\
\email{\{xiaofeihe,dengcai\}@cad.zju.edu.cn}
\and
University of Munich, Germany,
\email{jindong.gu@outlook.com}\\ 
\and
Tencent Data Platform, China,
\email{michaelzfli@tencent.com},\email{wl2223@columbia.edu}
}
\maketitle

\begin{abstract}
Vision Transformer (ViT), as a powerful alternative to Convolutional Neural Network (CNN), has received much attention. Recent work showed that ViTs are also vulnerable to adversarial examples like CNNs. To build robust ViTs, an intuitive way is to apply adversarial training since it has been shown as one of the most effective ways to accomplish robust CNNs. However, one major limitation of adversarial training is its heavy computational cost. The self-attention mechanism adopted by ViTs is a computationally intense operation whose expense increases quadratically with the number of input patches, making adversarial training on ViTs even more time-consuming. 
In this work, we first comprehensively study fast adversarial training on a variety of vision transformers and illustrate the relationship between the efficiency and robustness.
Then, to expediate adversarial training on ViTs, we propose an efficient Attention Guided Adversarial Training mechanism. Specifically, relying on the specialty of self-attention, we actively remove certain patch embeddings of each layer with an attention-guided dropping strategy during adversarial training. The slimmed self-attention modules accelerate the adversarial training on ViTs significantly. 
With only $65\%$ of the fast adversarial training time, we match the state-of-the-art results on the challenging ImageNet benchmark.

\end{abstract}

\section{Introduction}
\label{sec:intro}

Vision Transformers with the self-attention mechanism have been broadly studied and become de facto state-of-the-art models for many benchmarks. Recent works broadly investigated the traits of this new genre of architectures on computer vision tasks. Meanwhile, the adversarial robustness of Vision Transformers has also been intensively studied \cite{vit-rob-1,vit-rob-2,vit-rob-5,mahmood2021robustness,bai2021transformers,aldahdooh2021reveal,salman2021certified,yu2021mia,hu2021inheritance,mao2021rethinking,mao2021towards,mu2021defending,shi2021decision,joshi2021adversarial,fu2021patch}. To build robust Vision Transformers, an intuitive way is to apply adversarial training~\cite{shao2021adversarial,vit-rob-4} since it has been shown to be one of the most effective ways to achieve robust CNNs~\cite{fgsm,pgd,obfuscated}. 
However, one major limitation of adversarial training is its expensive computational cost. Adversarial training is known for requiring no extra cost during testing but greatly increasing the training cost. Huge efforts have been devoted to overcome this deficit~\cite{for-free,fast-free,vivek2020single,andriushchenko2020understanding,sriramanan2021towards,park2021reliably,Jia_2022_CVPR,jia2022boosting}. However, the philosophy of designing stronger ViTs has largely lifted up the computation intensity and weakened the performance of previously-proposed techniques. The specialty of the newly-proposed self-attention design~\cite{DBLP:conf/naacl/DevlinCLT19,DBLP:conf/nips/BrownMRSKDNSSAA20} in ViTs also introduces new challenges for accelerating adversarial training.

In this paper, we study the problem of how to efficiently carry out adversarial training on Vision Transformers. We first apply the state-of-the-art Fast Adversarial Training (Fast AT) algorithm~\cite{for-free,fast-free} on a variety of vision transformers and analyze how factors like attention mechanism, computational complexity, and parameter size influence the training quality. To the best of our knowledge, we are the first to accomplish a broad investigation on this topic. Our survey shows that, although ViTs outperform CNNs by a great margin on robustness, they have hugely increased the computational complexity. The self-attention mechanism adopted by ViTs is a computationally intense operation whose cost increases quadratically with the number of input patches. This newly-emerged module hampers the utilization of several techniques for accelerating adversarial training. Meanwhile, we find that large ViTs models also suffer from obvious catastrophic over-fitting problems~\cite{over-fitting}. This eventually leads to the degradation of robustness on ViTs with increasingly large capacity.

\begin{figure}[t!]
		\centering
\resizebox{0.98\textwidth}{!}{
		\subfigure[The illustration of layerwisely dropping patches]{
			\label{fig:drop-patch}
			\includegraphics[width=0.6\linewidth]{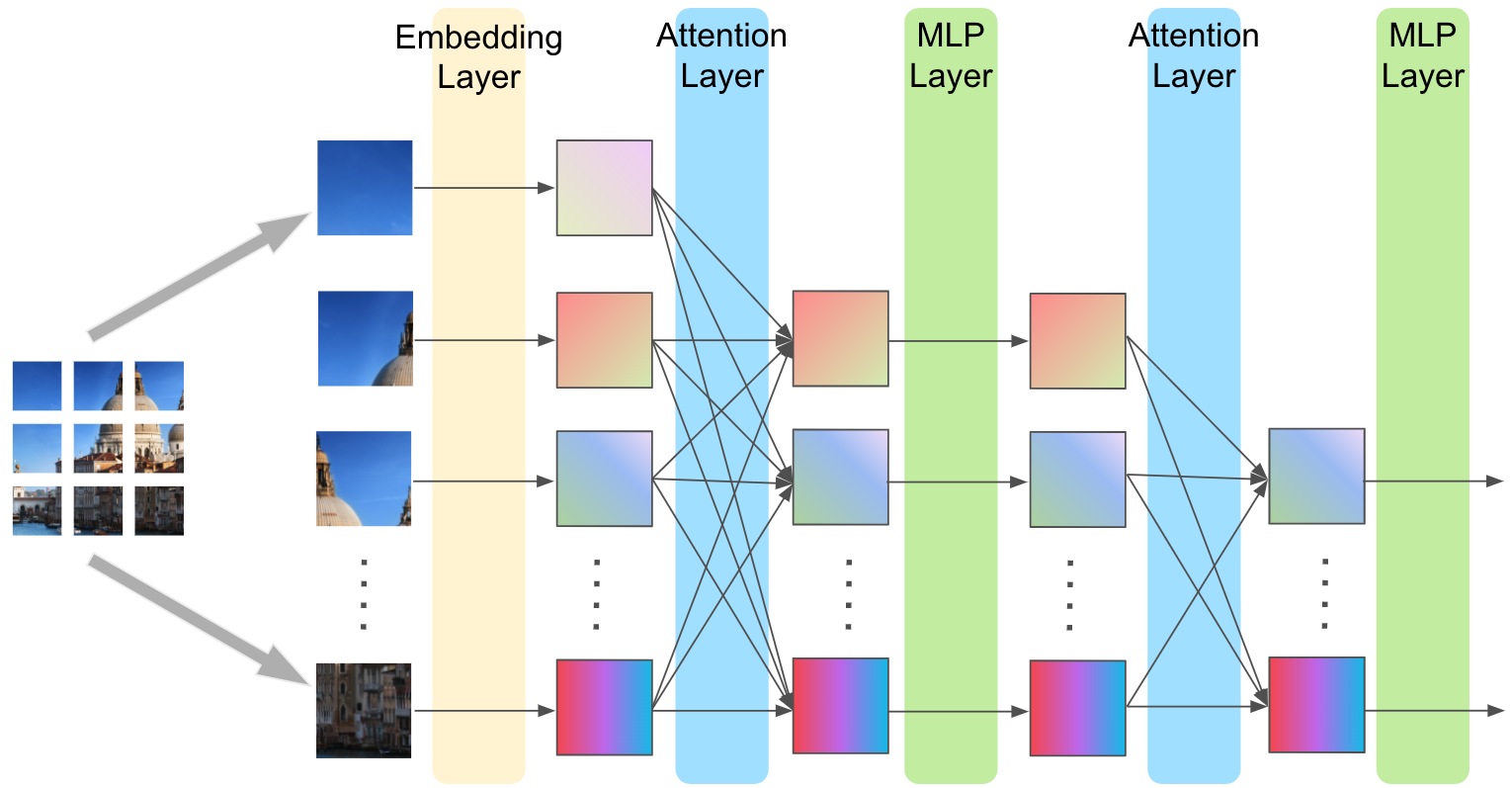}}
		\hspace{0.03\linewidth}
		\subfigure[AGAT performance]{
			\label{fig:performance}
			\includegraphics[width=0.33\linewidth]{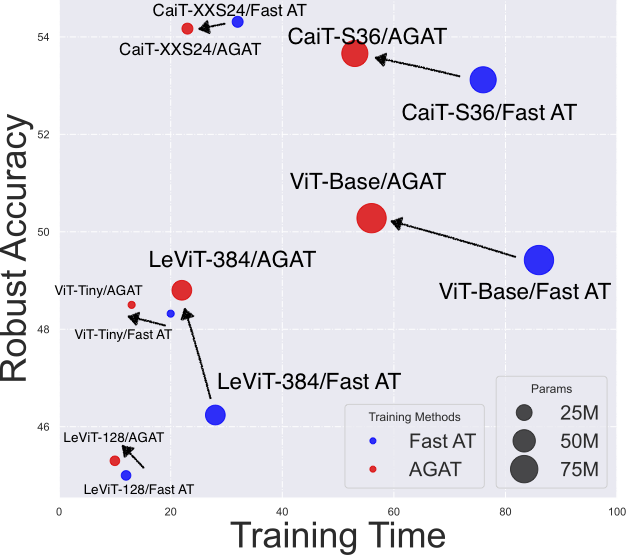}}
}
		\caption{(a) AGAT chooses to drop a certain proportion of image embeddings based on the attention information at each self-attention layer. (b) We plot the robust accuracy against the training time (hours) for various ViTs. AGAT substantially accelerates adversarial training, while maintaining or improving the robustness at the same time. }
		\label{fig:illustration}
\end{figure}

To make adversarial training efficient on the heavy-weight vision transformers, we investigate in accelerating adversarial training for vision transformers. Particularly, we leverage the specialty that the self-attention mechanism of transformers is capable of processing  variable-length inputs. This specialty of ViTs has been utilized in a wide range of applications, including processing variable-length word sequences for translations~\cite{DBLP:conf/naacl/DevlinCLT19}, mining graphs with unlimited edges~\cite{gat}, etc.
Recently, on vision tasks, several works~\cite{drop-patch-1,drop-patch-2,drop-patch-3} explored the possibility of dropping input image patches during training or testing for acceleration purposes. However, randomly dropping a certain number of input patches will inevitably hurt the training quality. Thus, many works have proposed adaptive designs~\cite{drop-patch-4,drop-patch-5} in the scenarios of a variety of targeted tasks.

Enlightened by the above works, we propose an Attention-Guided Adversarial Training (AGAT) mechanism, where we drop the patches based on the attention information. As illustrated by Fig.~\ref{fig:drop-patch}, our method intends to drop image embeddings after each layer of self-attention. Note that the self-attention layer is non-parametric and thus is not limited to a static number of inputs. We drop the embeddings with lower attention and keep the higher ones. Such a design will better preserve the feed-forward process and therefore guard the backward gradient computation of generating the adversarial examples.
As shown in Fig.~\ref{fig:performance}, AGAT gets to keep the training quality mostly unchanged or be improved by taking only $65\%$ training time. Our work matches the state-of-the-art results of adversarial robustness on the challenging benchmark of ImageNet.

\section{Relate Works}
\label{sec:related}

\textbf{Adversarial Training.}
Adversarial attacks~\cite{intrigue,fgsm,pgd,wei2019transferable,liang2020efficient,liang2021parallel} intend to endanger the performance of deep networks via repeatedly optimizing the input images with repect to the output of the model. To counter this unwanted deficit, various defensive approaches were proposed~\cite{meng2017magnet,dhillon2018stochastic,liao2018defense,xie2017mitigating,guo2017countering,song2017pixeldefend,samangouei2018defense,LiuJLC11}. Among these defenses, the methodology of adversarial training withstands most kinds of examinations and has become one of a few defenses that can consistently improve the robustness of deep networks when facing most attacks~\cite{obfuscated}. However, adversarial training is known to suffer from complexity issues~\cite{for-free,fast-free,llr}. Particularly, Fast AT~\cite{fast-free} enhances the single-step adversarial training with random initialization. Fast AT shows promising results on benchmark datasets. Later works~\cite{grad-align,ckpt,vivek2020single,andriushchenko2020understanding,sriramanan2021towards,park2021reliably} also proposed improved variants of Fast AT.

\noindent
\textbf{Vision Transformer.} The Transformer architecture and its self-attention mechanism were first proposed in the field of natural language processing (NLP)~\cite{DBLP:conf/nips/VaswaniSPUJGKP17,DBLP:conf/naacl/DevlinCLT19,DBLP:conf/nips/BrownMRSKDNSSAA20} and then adopted in the scenario of computer vision~\cite{DBLP:journals/corr/abs-2105-13677,DBLP:journals/corr/abs-2104-13840,WangJZYSL22}. After the huge efforts of a surge of explorations~\cite{vit,DBLP:conf/icml/TouvronCDMSJ21,DBLP:journals/corr/abs-2102-12122}, the Vision Transformer (ViT) has shown the potential to surpass the traditional convolutional neural networks. Then, researchers keep pushing this new philosophy of model design into a wide range of fields like high-resolution vision tasks~~\cite{swin,DBLP:journals/corr/crossformer}. Meanwhile, to reduce the huge computational expense that is brought by the densely modeled self-attention mechanism, various techniques have been proposed~\cite{DBLP:journals/corr/abs-2102-12122,DBLP:journals/corr/abs-2104-13840}. 



\noindent
\textbf{Adversarial Robustness of ViT.}
The adversarial robustness of ViT has also achieved great attention due to its impressive performance~\cite{bhojanapalli2021understanding,shao2021adversarial,paul2021vision,naseer2021intriguing,gu2021vision,bai2021are,benz2021adversarial,shi2021decision}. Some works~\cite{bhojanapalli2021understanding,shao2021adversarial,benz2021adversarial} first reported positive results where they showed that standard ViTs perform more robust than standard CNNs under adversarial attacks. The later works~\cite{bai2021are,gu2021vision} revealed that ViTs are not more robust than CNNs if both are trained in the same training framework. 
By adopting Transformers’ training recipes, CNNs can become as robust as Transformers on defending against adversarial attacks. In both sides, we can observe that the clean accuracy of standard models can be easily reduced to near zero under standard attack protocols. In addtion, Fu et al.~\cite{fu2021patch} studied attacking ViTs in a patch-wise approach, which reveals the unique vulnerability of ViTs. To boost the adversarial robustness of ViTs, recent works~\cite{tang2021robustart,DBLP:journals/corr/abs-2204-00993} explored multiple-step adversarial training to ViTs. 
Shao et al.~\cite{shao2021adversarial}, tested the vanilla adversarial training on CIFAR10.
However, multi-step adversarial training is computationally expensive. And in this work, we take the step of exploring fast single-step adversarial training on ViT models.

\section{Fast Adversarial Training on Vision Transformers}

We first comprehensively study the Fast AT~\cite{fast-free} algorithm on vision transformers. Fast AT is designed to be efficient so that it can be applied to large models (e.g., ResNet-101~\cite{resnetv1}) on large-scale datasets (e.g., ImageNet~\cite{DBLP:journals/ijcv/RussakovskyDSKS15}). Specifically, Fast AT refines the standard single-step FGSM~\cite{fgsm} algorithm by adopting a large random perturbation as the starting point for searching adversarial examples. By doing so, Fast AT is supposed to effectively resist the catastrophic overfitting~\cite{fast-free} of training with the plain FGSM. Thus, Fast AT preserves the effectiveness while significantly improves the efficiency over the multi-step PGD~\cite{pgd} algorithm. To better understand the robustness of the newly-developed ViT models, in this section, we apply Fast AT to a wide range of ViTs.

We select nineteen models of different sizes from five vision transformer families, including ViT~\cite{vit}, CaiT~\cite{cait}, LeViT~\cite{levit}, SwinTransformer~\cite{swin}, and CrossFormer~\cite{DBLP:journals/corr/crossformer}. The selected models cover a wide range of model designs, including hybrid models~\cite{levit}, slim models~\cite{levit,DBLP:journals/corr/crossformer}, constrained attention~\cite{swin,DBLP:journals/corr/crossformer}, and multi-scale attention~\cite{levit,DBLP:journals/corr/crossformer}. Following the settings of Fast AT~\cite{fast-free}, we set the perturbation radius to $2/255$ for ImageNet and test the adversarially trained models with the 100-step PGD attack. Our training schedule aligns with SwinTransformer~\cite{swin} and DeiT~\cite{deit}. We keep all hyper-parameters, \eg, image size, training epoch, and data augmentation, identical for all models. We also present the results of Fast AT on the CNN models of ResNet-50 and ResNet-101 for comparison. As shown by Fig.~\ref{fig:2a}, ViTs are consistently more robust than CNNs. This aligns with concurrent researches on model robustness~\cite{robustart}. We conclude draw novel observations as follows:

\begin{figure}[t!]
		\centering
		\subfigure[Robust Accuracy for Different Attacking Steps. ]{
		\label{fig:2a}
			\includegraphics[width=0.64\linewidth]{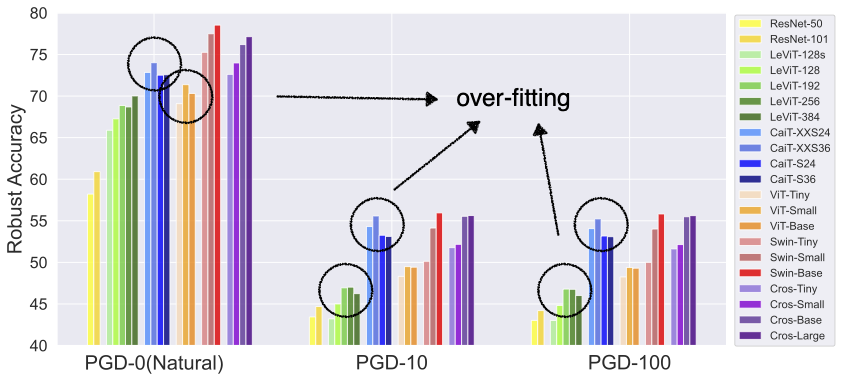}}
		\subfigure[Training Curve]{
		\label{fig:2b}
			\includegraphics[width=0.29\linewidth]{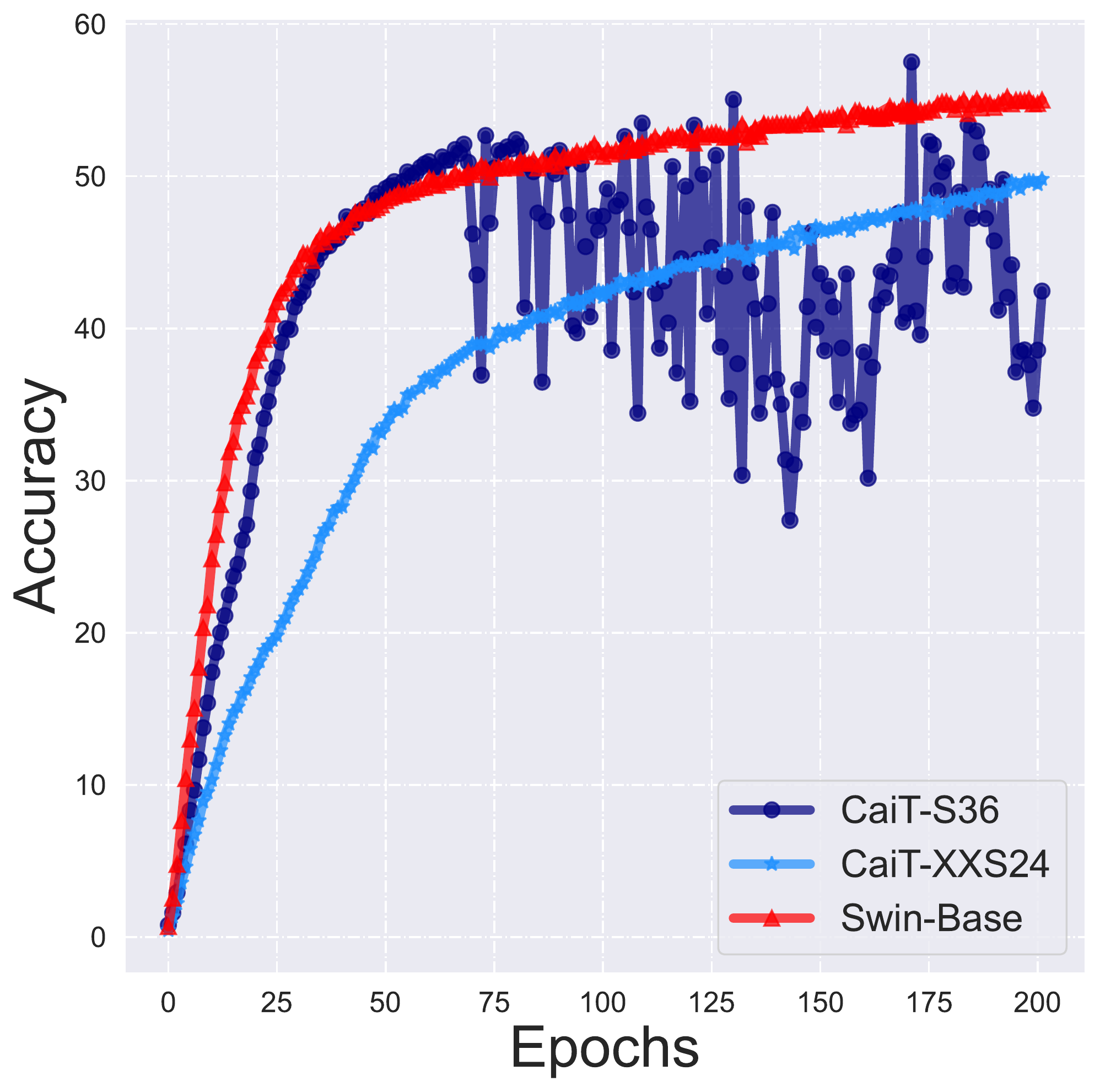}} 
		\\
		\subfigure[Performance Outliers ]{
		\label{fig:2c}
			\includegraphics[width=0.31\linewidth]{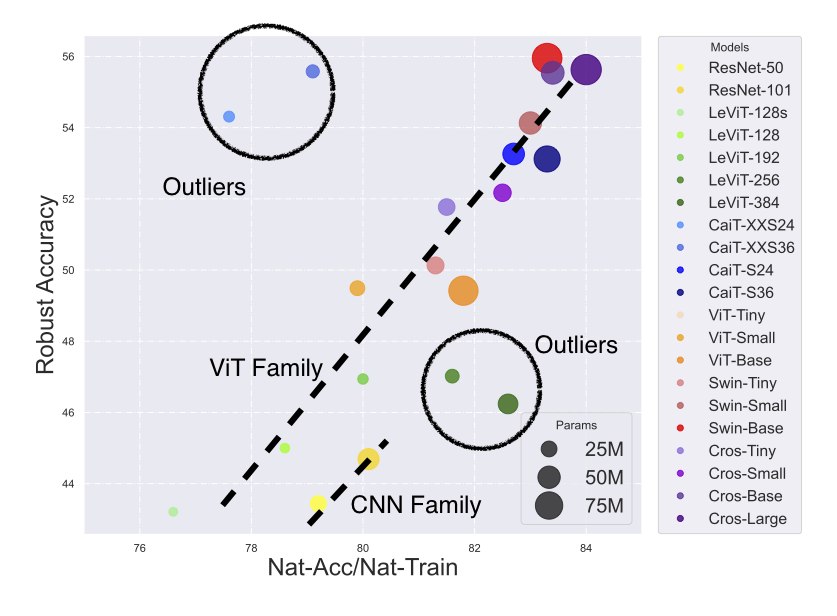}}
		\subfigure[Performance Alignment ]{
		\label{fig:2d}
			\includegraphics[width=0.31\linewidth]{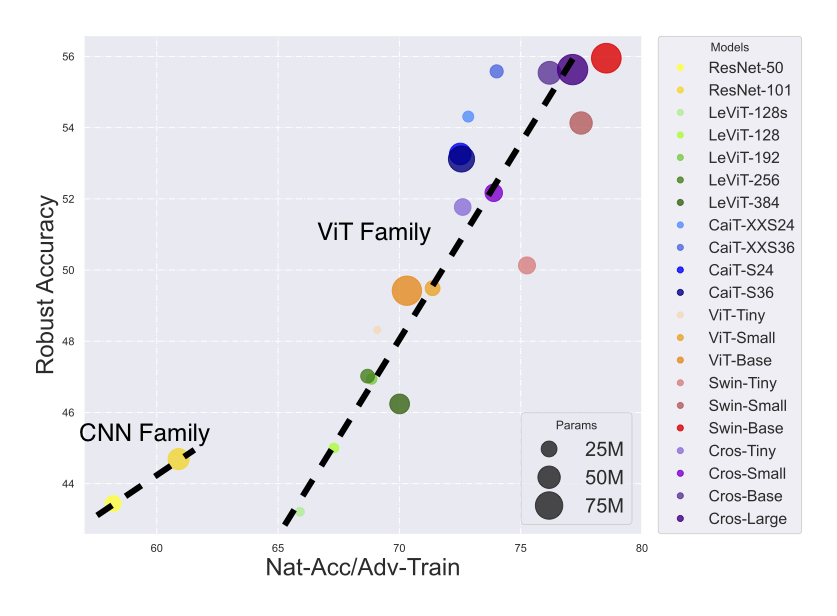}}
		\subfigure[Computation Intensity]{
		\label{fig:2e}
			\includegraphics[width=0.31\linewidth]{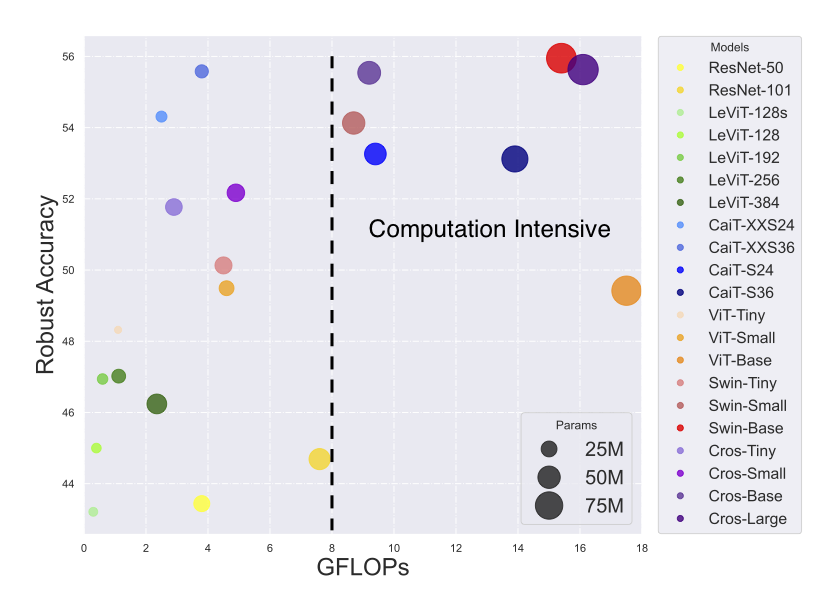}}
		\caption{(a) Robust accuracy for various ViTs. (b) Large ViTs like CaiT-S36 may suffer from obvious unstable training and over-fitting, while designs like window attention~\cite{swin} can alleviate the issue. (c) The perfromance of ViTs on Fast AT may not align with their natural accuracy of natural training. (d) In contrast, the natural and robust accuracies of ViTs align with each other when both are under Fast AT. (e) Large ViTs greatly increase the computational complexity. }
\end{figure}

\begin{enumerate}

\item \textbf{Within the same transformer family, larger transformers do not always result in better robustness.} In Fig.~\ref{fig:2a}, the transformer families of LeViT, CaiT, and ViT exhibit a pattern of over-fitting. Namely, as the models get larger, the network robustness learned by Fast AT degrades conversely. For instance, Cait-S36 performs worse than CaiT-XXS24. Fig.~\ref{fig:2b} provides more details of the above over-fitting issue. The optimization of the CaiT-S36 model gradually degrades after a certain point. In contrast, the CaiT-XXS24 model possesses a monotonically increasing  training curve. This aligns with previous findings that ViTs may suffer from more severe over-fitting on natural tasks and need the assistance of aggressive data augmentation schedules~\cite{deit}. Moreover, transformers with a constrained attention mechanism can alleviate the over-fitting problem. In Fig.~\ref{fig:2b}, unlike CaiT-S36, the large model of Swin-Base shows a steady training curve.

\item 
\textbf{Among different transformer families, the attention mechanism designed for better natural performance not necessarily results in better robustness.} In Fig.~\ref{fig:2c}, for each transformer architecture, we plot its natural performance under natural training against its robust performance under Fast AT. Among different transformer architectures, the two metrics of natural accuracy and robust accuracy approximately form a line, indicating the close relation between natural and robust performances. Compared with the line formed by CNNs, vision transformers consistently achieve better robustness on models with similar natural performance. However, we can observe a few outliers of models from the LeViT and CaiT families. Specifically, the hybrid design of LeViT can achieve high robustness with very small models. Large CaiTs, despite being effective on the natural task, result in obvious inferiority on robustness. Notice that, for each point in Fig.~\ref{fig:2c}, the two metrics of the vertical axis and the horizontal axis are evaluated on two models, either adversarially-trained or naturally-trained, of the same architecture. When we plot the natural accuracy and robust accuracy, both of which are under Fast AT, as shown by Fig.~\ref{fig:2d}, the two metrics consistently align with each other without any outlier, revealing the difference between Fast AT and standard training.

\item
\textbf{SOTA ViTs suffer from a severe efficiency issue and
require much more training time than SOTA CNNs.}
The above over-fitting problem mainly shows on large ViTs like Cait-S36 or LeViT-384, but not on the small ones. This is reasonable since models like Cait-S36 are consistently larger than commonly-used CNNs. However, the over-fitting is not the only problem of adopting larger and larger ViTs. These increasingly large models have hugely lifted up the computation intensity. In Fig.~\ref{fig:2e}, we plot robustness (Robust Accuracy) against computation intensity (GFLOPs). State-of-the-art ViTs can be a few times larger than CNNs. This makes utilizing Fast AT even harder since the self-attention module is more intricate to accelerate.

\end{enumerate}

\section{Efficient Adversarial Training on Vision Transformers}
\label{sec:connect}

As discussed in the last section, one major problem that hinders the deployment of adversarial training on vision transformers is the efficiency issue. Adversarial training is known to be computationally intensive. This greatly hampers its usage on large-scale models or datasets. Various techniques have been proposed to mitigate this problem. However, with the revolution brought by vision transformers, many existing techniques such as variable-resolution training~\cite{fast-free} have been unusable. More importantly, there is a rising trend of adopting increasingly large ViTs for better performance. The complexity of these enormous ViTs is too large to afford, even for efficient algorithms like Fast AT.

In this section, we first analyze the computational complexity of popular ViTs. Our analysis shows that ViT requires a much longer time to finish adversarial training, which is caused by the large computational cost of ViT brought by a large number of input patches. Then, we explore the input patches to reduce the brought computational cost. Given the fact that the flexibility of self-attention allows ViTs to process an arbitrary length of image patches, we explore a random patch dropping strategy to reduce the computation. The dropping operation with the reduced number of patches can accelerate adversarial training, as expected. However, the naive dropping strategy will also hurt robustness. To address the above issues, we propose our Attention-Guided Adversarial Training algorithm, which selectively drops patches based on attention magnitude.

\subsection{Computation Intensity of ViTs}

We first formally formulate our task. For each matrix, we present its shape in the lower right corner and its index in the upper right corner. Denote the input feature as $\Xb_{p \times d}$, which consists of a sequence of $p$ embeddings with the dimension being $d$ : $\Xb=[X^1_{d}, X^2_{d}, ..., X^p_d]$. Each embedding relates to a specific non-overlapped patch of the input image. Vision transformer consists of a list of blocks, each of which consists of two kinds of computation, i.e., the Multi-head Self-Attention layer (MSA) and the Multi-Layer Perceptron layer (MLP). In the MSA module, $\Xb$ is first normalized via Layer Normalization and then transformed to the \textit{query, key, and value} matrices ($\Kb,\ \Qb,\ \Vb$).
\begin{align}
[\Kb_{p\times d},\ \Qb_{p\times d},\ \Vb_{p\times d} ] = \text{LayerNorm}(\Xb_{p\times d})\Wb^{1}_{d\times 3d} .
\end{align}
For the multi-head design, we partition the $\Kb,\ \Qb,\ \Vb$ matrices of shape $p\times d$ into $h$ heads, with each part having a shape of $p\times \frac{d}{h}$. Then, taking the first head as an example, $\Vb^1$ will be re-weighted by $\Ab^1$ with the following form:
\begin{align}
\text{Attn}(\Kb^1_{p\times \frac{d}{h}}, \Qb^1_{p\times \frac{d}{h}}, \Vb^1_{p\times \frac{d}{h}})
=
\text{SoftMax}(\Qb^1{\Kb^1}^{\top} / \sqrt{d} + \Bb)\Vb^1
=
\Ab^1_{p\times p}\Vb^1 .
\end{align}
$\Bb$ is a learnable bias.
Note that all the column vectors of $\Ab^1$ are normalized by \text{Softmax} and thus have a summation of $1$ in each. Then,  $\Ab\Vb$ value of each head will be concatenated and transformed to the output of MSA.
\begin{align}
\Xb'_{p\times d} = \text{Concat}(\Ab^1\Vb^1, \Ab^2\Vb^2,..., \Ab^h\Vb^h)_{p\times d} \Wb^2_{d\times d} .
\label{eq:msa-w2}
\end{align}
A following MLP module will take in the output of MSA, $\Xb'$, and transform each embedding with Layer Normalization and GELU activation.
\begin{align}
\Xb''_{p \times c} = 
\text{GELU}\Big[
\text{LayerNorm}(\Xb'_{p \times d})\Wb^{3}_{d\times 4d}
\Big]
\Wb^{4}_{4d\times d} .
\end{align}
The computational complexity of the above process is:
\begin{align}
\mathrm{\Omega}(\text{MSA}) = 4pd^2+2p^2d+pd ;
\quad
\mathrm{\Omega}(\text{MLP}) = 8pd^2+pd .
\end{align}
A typical vision transformer will consecutively conduct the above process to generate the final image representation for prediction.
Take the ViT-Base model as an example. Each layer has $d=768$. Therefore, we have $\mathrm{\Omega}(\text{MSA})+\mathrm{\Omega}(\text{MLP})=7\times10^6p+1.5\times10^3p^2$. Since $7\times10^6\gg 1.5\times10^3$ and $p$ is mostly around $2\times10^2$, the computational complexity of the entire ViT is approximately linear to $p$.

\subsection{Dropping Patch: The Flexibility of Self-Attention}

Different from the convolutional operation where the hyperparameters (e.g., kernel size, padding size) are supposed to be fixed, the self-attention operation does not require the inputs with fixed length. For instance, this flexibility of self-attention is leveraged to process an arbitrary length of words in NLP tasks. Similarly, the flexibility makes its adaption to graph data feasible, in which different nodes can have a different number of connected edges~\cite{gat}. When transformers with self-attention mechanisms have been introduced into computer vision tasks, researchers also investigate dynamically dropping the patches or the embeddings in the forward pass of a ViT model~\cite{drop-patch-1,drop-patch-2,drop-patch-3}. It is found that, when a constrained quantity of patches are dropped, the forward inference can be significantly accelerated. Meanwhile, the performance of the model will be only slightly degraded~\cite{drop-patch-4,drop-patch-5}. Several works utilize this feature to design new mechanisms for their own unique purposes.

\begin{figure}[t!]
		\centering
		\subfigure[Baselines]{
		\label{fig:3a}
			\includegraphics[width=0.3\linewidth]{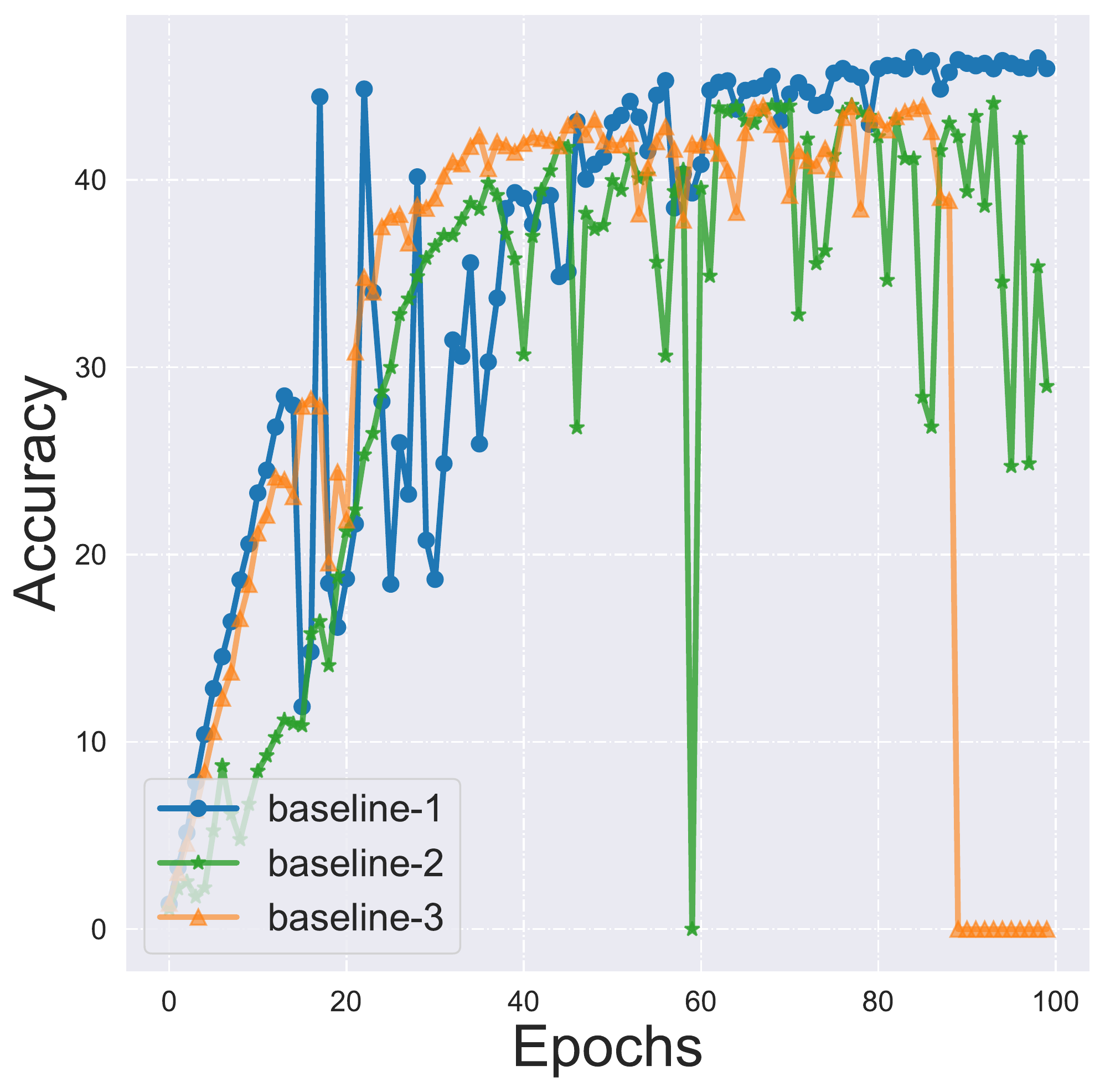}}
		\subfigure[Dropping Patches]{
		\label{fig:3b}
			\includegraphics[width=0.3\linewidth]{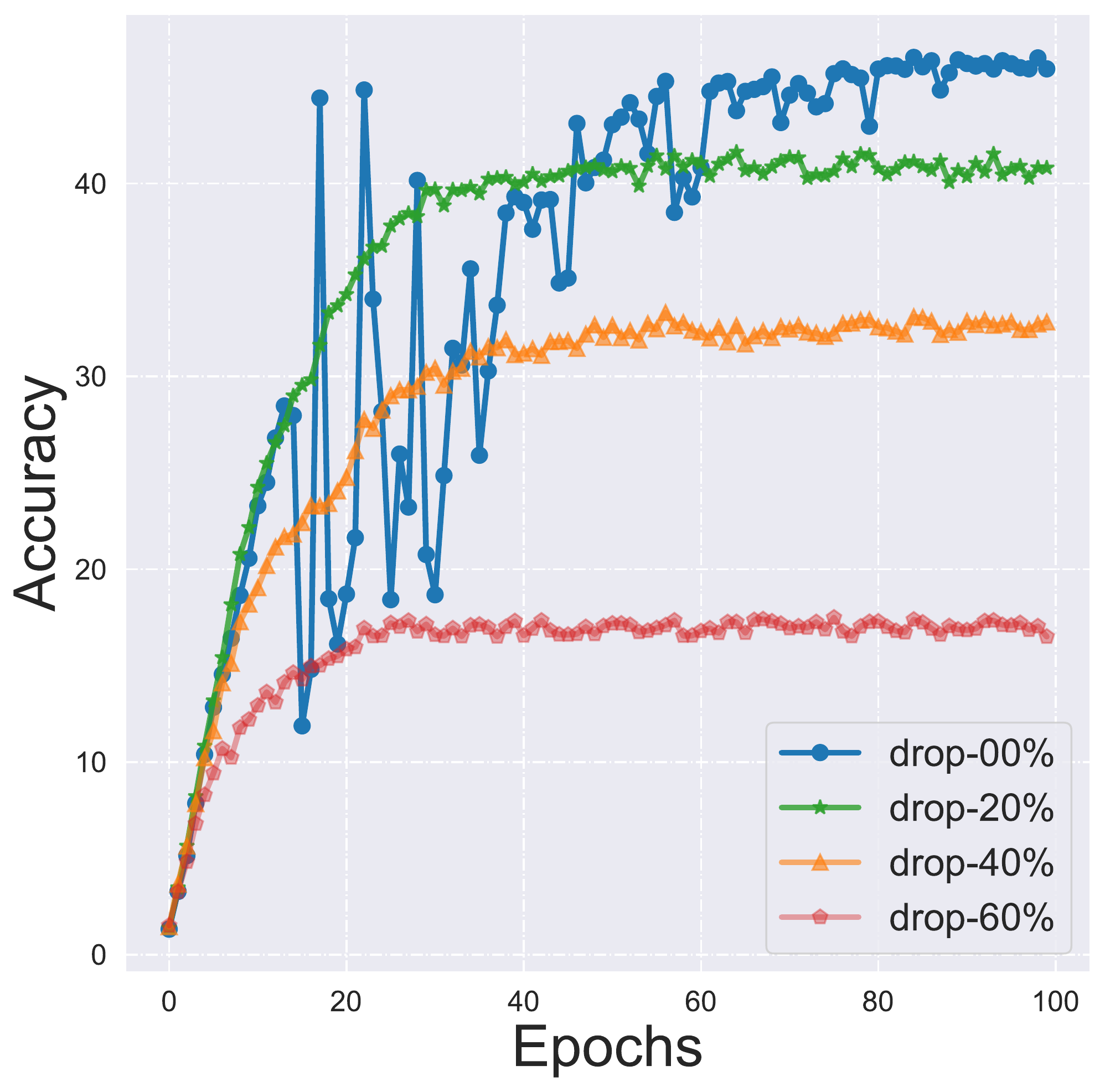}}
		\subfigure[Attention Dropout]{
		\label{fig:3c}
			\includegraphics[width=0.3\linewidth]{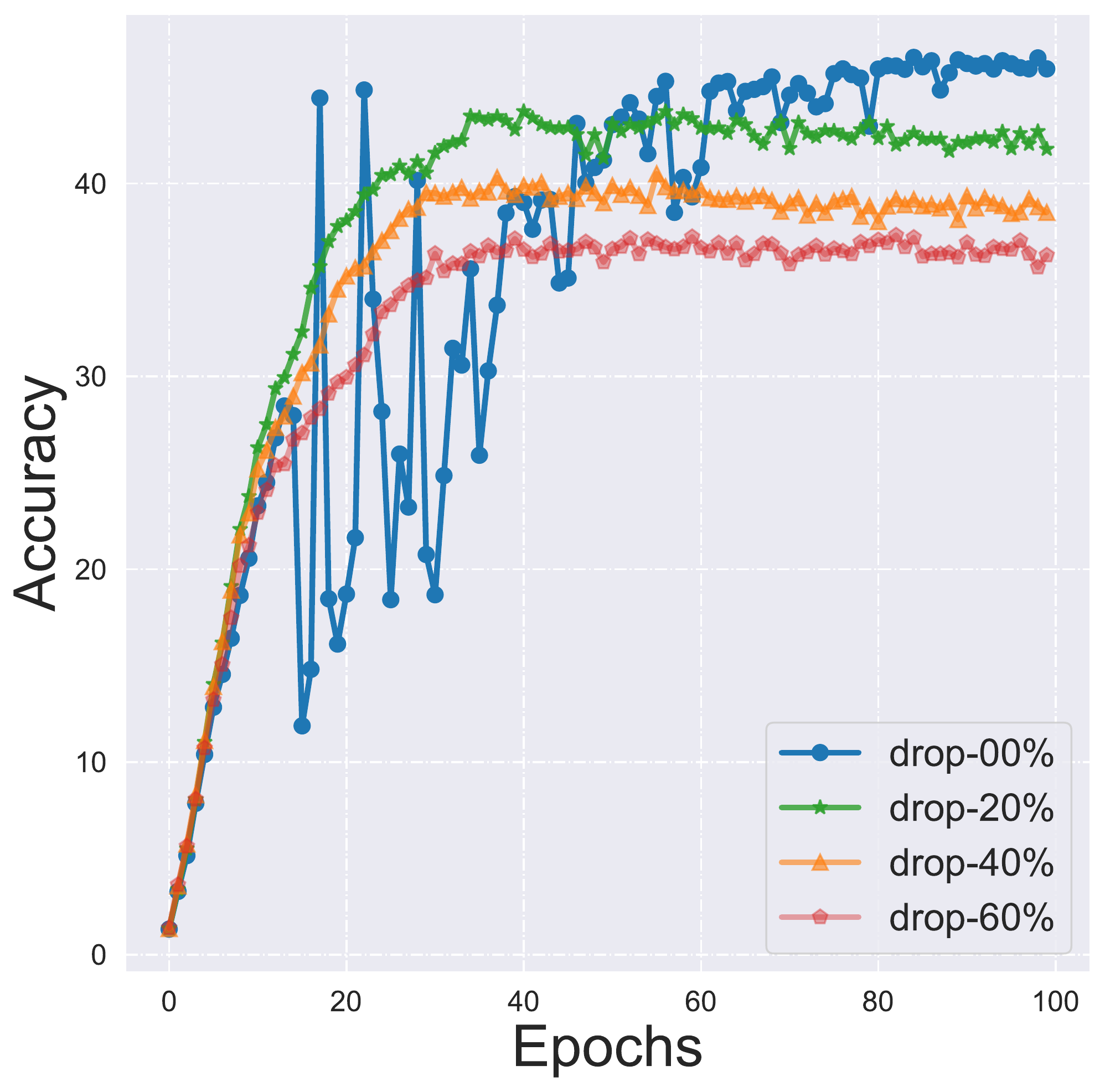}}
		\caption{(a) We repeat the baseline without any dropping for three times. We can observe the training is very unstable. (b) Training curve when we drop various rates of input patches. (c)Training curve when we adopt various rates of attention dropout.}
\end{figure}

In this work, we also explore a patch dropping strategy to accelerate adversarial training given the excellent trade-off it achieves. We first test the scheme of randomly dropping a certain number of input patches to see how it influences the training quality of Fast AT. We report the results in Fig~\ref{fig:3b}, where we plot the robustness against the training epoch for different ratios of dropping. Note that no patches will be dropped during inference in the testing stage. When the number of input patches is reduced, the forward inference of ViT can be accelerated. Surprisingly, from the figure, we also observe that the dropping operation also stabilizes the adversarial training and alleviates the phenomenon of catastrophic over-fitting~\cite{fast-free,fat-over-1,fat-over-2}. As shown by Fig~\ref{fig:3a}, we repeat Fast AT without any dropping for three times. The training procedure can be very unstable and occasionally drop to zero accuracy. We conjecture that it is the regularization effect brought by the patch dropping operation that stabilizes Fast AT. To further verify this conjecture, we test ViTs equipped with the dropout operation as in DeiT~\cite{deit}. The dropout module is applied right after the self-attention module. As shown in Fig~\ref{fig:3c}, like dropping patches, the attention dropout module also stabilizes Fast AT. Unlike dropping patches, the  dropout module cannot save computation. However, the final robust accuracy can be reduced in both cases when dropping is applied. The random patch dropping strategy poses a dilemma. Namely, it brings both acceleration and performance degradation. In the following section, we will present our attention-guided patch dropping strategy, where we achieve a better trade-off between efficiency and effectiveness.

\begin{figure}[t!]
		\centering
\resizebox{0.98\textwidth}{!}{
		\subfigure[Plain Self-Attention.]{
			\includegraphics[width=0.45\linewidth]{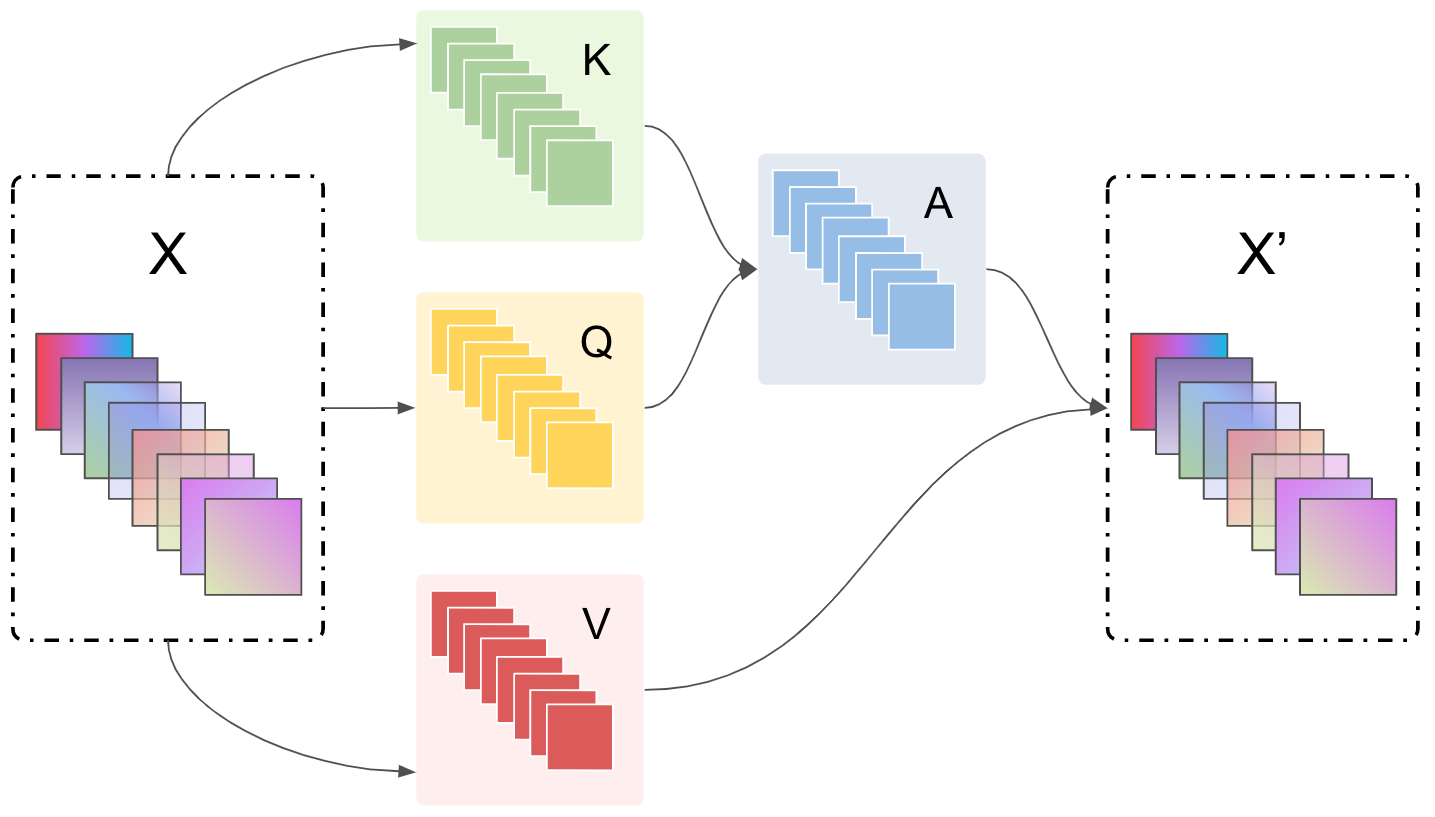}}
		\hspace{0.01\linewidth}
		\rulesep
	    \hspace{0.01\linewidth}
		\subfigure[Attention of AGAT.]{
			\includegraphics[width=0.537\linewidth]{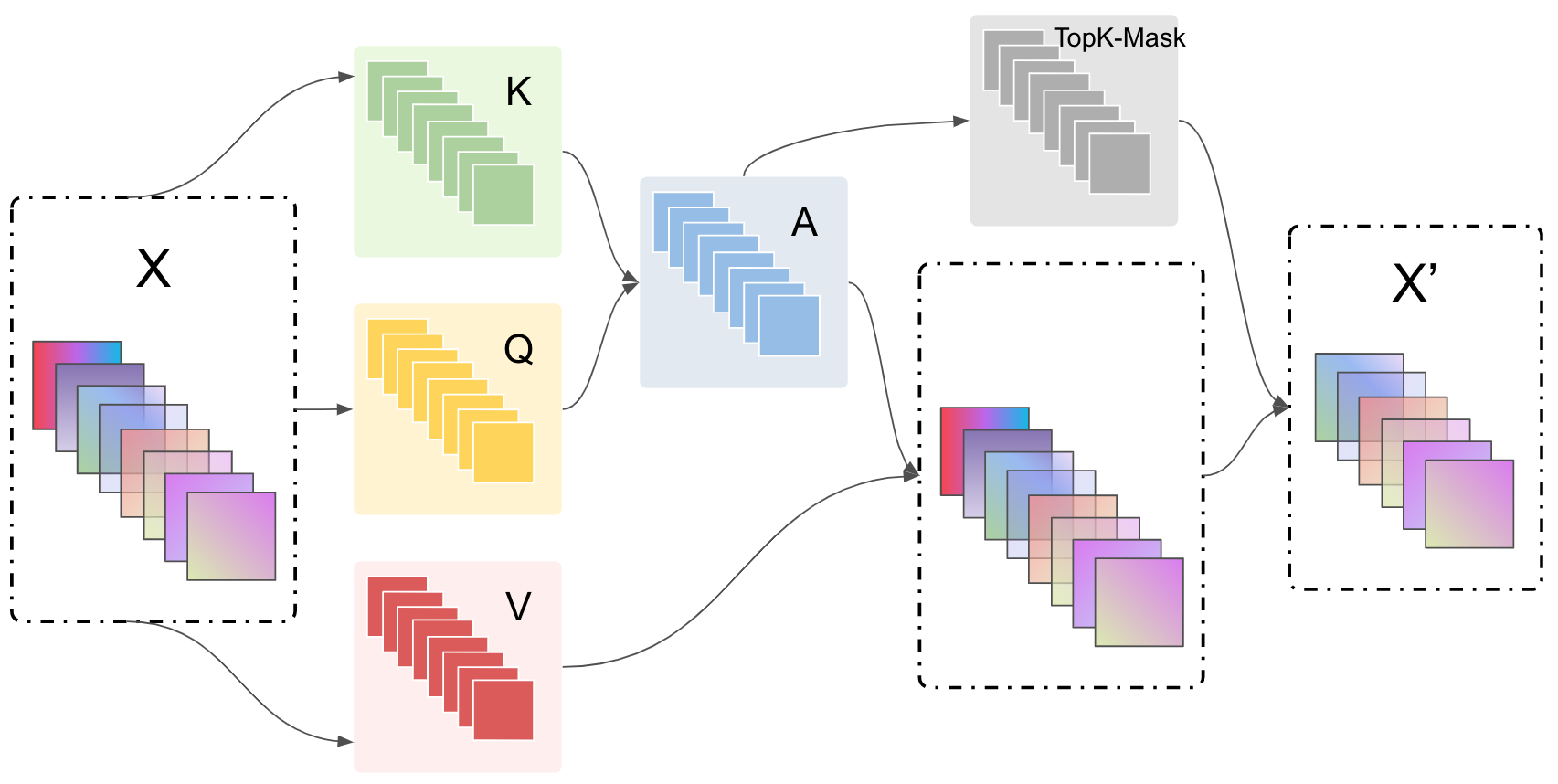}}
}
		\caption{The illustration of the plain self-attention layer and the attention layer of AGAT. The embedding mask is directly generated from $\Ab$ and is non-parametric. Thus, for the same model, one can adopt our AGAT during training and the plain self-attention during testing, respectively.}
\end{figure}

\subsection{Attention-Guided Adversarial Training}

It is known that adversarial training utilizes adversarial attacks to generate examples so that the network can learn to fit the generated adversarial examples. This is a typical hard example mining framework. The more powerful the adversarial examples are, the more robust the learned network will be. The quality of the adversarial examples relies on the adversarial attacking algorithm. And the attack algorithm depends on accurately estimating the gradient of the input pixels with respect to the loss function. Thus, a major insight of achieving our goal is to drop the patches that will barely hamper the gradient estimation. A similar philosophy has been utilized for sparse attacks or black-box attacks.

Recent works studied the learned attention and found that the magnitude of the attention can reveal how salient an embedding is~\cite{CaronTMJMBJ21}. This enlightens us that we can utilize this ready-made information to filter salient embeddings. Thus, we sum $\frac{1}{h}\sum_{i=1}^{h} \Ab^i$ by row and generate an index vector $\ab$. Note that the column of $\frac{1}{h}\sum_{i=1}^{h} \Ab^i$ is the weighted-average parameter and thus always equal to $1$. It indicates how much a generated embedding receives the information of each input embedding. In contrast, the row of $\frac{1}{h}\sum_{i=1}^{h} \Ab^i$ reveals how much each input embedding influences the output embeddings. This value differs from embedding to embedding. Thus, we choose to select the top-k embeddings based on their magnitude in $\ab$. Then, the formulation of \eqref{eq:msa-w2} becomes:
\begin{align}
\Xb'_{k\times d} = 
\text{MaskBy}\Big[
\text{Concat}(\Ab^1\Vb^1, ..., \Ab^h\Vb^h),
\text{Topk}( \ab)
\Big]_{k\times d}
\Wb^2_{d\times d} .
\end{align}
We drop the embeddings after the weighted average calculation of $\Ab\Vb$ so that the magnitude of embeddings will be kept stable. The number of embeddings will reduce from $p$ to $k$. To fully utilize the attention information in each layer, we propose a layer-wise exponential dropping scheme. Namely, in each layer, we drop a constant proportion of patches. Thus, this scheme will drop more embeddings on deeper layers, where the embeddings are consistently more redundant~\cite{drop-patch-2}. We set the dropping rate to $0.9$. On a 12-layer ViT-Base model, the final layer will process only $31\%$ number of embeddings and save more than $40\%$ FLOPs of the entire model. A detailed implementation of our Attention-Guided Adversarial Training is shown in Algorithm~\ref{alg:agat}. Our AGAT only modifies the training process. During testing, we use the original model for prediction. The class token will not be dropped when it involves the feed-forward procedure.

\begin{algorithm}[t]
\caption{{AGAT attention code (PyTorch-like)}}
\label{alg:agat}
\definecolor{codeblue}{rgb}{0.25,0.5,0.25}
\lstset{
	backgroundcolor=\color{white},
	basicstyle=\fontsize{7.2pt}{7.2pt}\ttfamily\selectfont,
	columns=fullflexible,
	breaklines=true,
	captionpos=b,
	commentstyle=\fontsize{7.2pt}{7.2pt}\color{codeblue},
	keywordstyle=\fontsize{7.2pt}{7.2pt},
}
\begin{lstlisting}[language=python]

def init(num_heads, dim, drop_rate): 
    head_dim = dim // num_heads # num_heads: the number of heads, dim: embedding dimension
    scale = head_dim ** -0.5

    qkv = nn.Linear(dim, dim * 3)
    proj = nn.Linear(dim, dim)

def forward(self, x): # x: input tensor with the shape of (b, p, d);
    b, p, d = x.shape # b: batch size; p: patch number
    q, k, v = qkv(x).reshape(b, p, 3, num_heads, head_dim).permute(2, 0, 3, 1, 4).unbind(0)  

    attn = (q @ k.transpose(-2, -1)) * scale
    attn = attn.softmax(dim=-1) # b num_heads p p 
    
    own_attn = torch.sum(torch.sum(attn, dim=1), dim=-2) # b p
    kept_num = int(p * drop_rate) - 1 # compute the number of kept embeddings
    _, rank_indices = torch.topk(own_attn[:,1:], k=kept_num, dim=-1) # b k
    rank_indices = rank_indices.unsqueeze(-1).repeat(1,1,dim) # for the API of torch.gather 
 
    x = (attn @ v).transpose(1, 2).reshape(B, N, C)
    x = torch.gather(x, dim=1, index=rank_indices) # drop embeddings

    return proj(x)
        
\end{lstlisting}
\end{algorithm}

\section{Experiments}

\subsection{Experimental Setup}

\noindent\textbf{Dataset} We evaluate our method on the challenging ImageNet~\cite{imagenet} dataset. Due to the huge computational expense of adversarial training, most adversarial training approaches are only verified on relatively small datasets such as CIFAR10~\cite{cifar} or MNIST~\cite{mnist}. Our efforts on improving the efficiency of adversarial training allow us to apply adversarial training to large-scale datasets. The input image size is $224$ on all models for a fair comparison. 

\noindent\textbf{Training Schedule} Following previous works~\cite{deit,DBLP:journals/corr/crossformer}, we use the  AdamW~\cite{adamw} optimizer for training for 300 epochs with a cosine decay learning rate schedule. The initial learning rate is set to $0.001$. The first 20 epochs adopt the linear warm-up strategy. The batch size is 1024 split on 8 NVIDIA A100 GPUs. 

\noindent\textbf{Evaluation Metrics} We report our major results on two metrics, robust accuracy and GFLOPs. We mainly focus on improving the speed of training and keeping the learned robustness unchanged at the mean time. For a direct impression of training speed, we also record the training time for each method. Note that the training time is not only determined by the efficiency of the training algorithm, but also the IO speed and many other nonnegligible factors.

\noindent\textbf{Adversarial Attack} We choose the powerful multi-step PGD attack~\cite{pgd} with the perturbation radius being $2/255$ or $4/255$ and the optimization step being $20$ or $100$~\cite{xie2017mitigating}. We also test different kinds of attacks, including black-box attacks, to rule out the possibility of obfuscated gradient~\cite{obfuscated}.

\begin{table*}[t!]
	\caption{Adversarial Training on The ImageNet Dataset. In most cases, our Attention-Guided Adversarial Training on ViTs achieves comparable clean performance and robust accuracy to Fast-AT with much less time. The conclusion still holds when different perturbation ranges are applied.}
	\label{table:main}
	\centering
	\resizebox{0.98\textwidth}{!}{
	\begin{tabular}{ ccccccccccccc } 
		\toprule
		& & & & & &
		& \multicolumn{3}{c}{$\epsilon=2/255$}
		& \multicolumn{3}{c}{$\epsilon=4/255$}\\
		 \cmidrule(r){8-10}
		 \cmidrule(r){11-13}
		Model 
		& Params 
		& \tabincell{c}{Block\\Number}
		& \tabincell{c}{Training\\Method}
		& \tabincell{c}{Dropping\\Rate}
		& FLOPs
		& \tabincell{c}{Training\\Time}
		& \tabincell{c}{Nat-\\Acc.}
		& PGD-20
		& PGD-100
		& \tabincell{c}{Nat-\\Acc.}
		& PGD-20
		& PGD-100 \\
		\midrule
		\multirow{3}{*}{ResNet-50~\cite{resnetv1}} 
		& \multirow{3}{*}{25.6M} 
		& \multirow{3}{*}{16} 
		& Free AT~\cite{for-free}
		& -
		& 3.8G
		& 46H
		& \textbf{62.28} & \textbf{43.77} & \textbf{43.44}
		& \textbf{58.31} & 30.89 & 30.71  \\ 
		& 
		&
		& Fast AT~\cite{fast-free}
		& -
		& 3.8G
		& \textbf{14H}
		& 58.20 & 43.62 & 43.31
		& 52.62 & 30.17 & 30.13  \\ 
		& 
		&
		& Grad Align~\cite{grad-align}
		& -
		& 3.8G
		& \textbf{14H}
		& 57.44 & 42.61 & 42.46
		& 53.66 & \textbf{31.18} & \textbf{31.01}  \\ 
		\midrule
		\multirow{3}{*}{ResNet-101~\cite{resnetv1}} 
		& \multirow{3}{*}{44.5M} 
		& \multirow{3}{*}{33} 
		& Free AT~\cite{for-free}
		& -
		& 7.6G
		& 51H
		& \textbf{64.37} & 43.27 & 43.14
		& \textbf{60.41} & 31.17 & 31.14  \\
		& 
		&
		& Fast AT~\cite{fast-free}
		& -
		& 7.6G
		& \textbf{17H}
		& 60.90 & \textbf{44.57} & \textbf{44.11}
		& 55.62 & \textbf{33.02} & \textbf{33.26}  \\ 
		& 
		&
		& Grad Align~\cite{grad-align}
		& -
		& 7.6G
		& \textbf{17H}
		& 60.12 & 43.28 & 43.04
		& 53.94 & 30.27 & 30.21  \\ 
		\midrule
		\multirow{3}{*}{CaiT-XXS24~\cite{cait}} 
		& \multirow{3}{*}{12.0M} 
		& \multirow{3}{*}{24}
		& Fast AT~\cite{fast-free}
		& 0.0
		& 2.5G
		& 32H
		& \textbf{72.84} & \textbf{54.31} & \textbf{54.26}
		& \textbf{68.77} & \textbf{32.14} & \textbf{32.09}  \\ 
		& 
		&
		& Random
		& 0.4
		& \textbf{1.5G}
		& \textbf{22H}
		& 65.62 & 39.81 & 39.52
		& 60.87 & 31.54 & 31.46  \\ 
		& 
		&
		& AGAT
		& 0.05
		& \textbf{1.5G}
		& 23H
		& 71.15 & 54.17 & 54.08
		& 68.22 & 31.46 & 31.39  \\ 
		\midrule
		\multirow{3}{*}{CaiT-XXS36~\cite{cait}} 
		& \multirow{3}{*}{17.3M} 
		& \multirow{3}{*}{36}
		& Fast AT~\cite{fast-free}
		& 0.0
		& 3.8G
		& 38H
		& \textbf{74.01} & 55.58 & 55.41
		& 73.54 & 34.61 & 34.28  \\ 
		&
		& 
		& Random
		& 0.4
		& \textbf{2.7G}
		& \textbf{25H}
		& 69.25 & 50.13 & 49.89
		& 65.12 & 28.00 & 27.91  \\ 
		&
		& 
		& AGAT
		& 0.03
		& \textbf{2.7G}
		& \textbf{25H}
		& 73.83 & \textbf{55.81} & \textbf{55.72}
		& \textbf{73.91} & \textbf{35.22} & \textbf{35.19}  \\ 
		\midrule
		\multirow{3}{*}{CaiT-S36~\cite{cait}} 
		& \multirow{3}{*}{68.2M} 
		& \multirow{3}{*}{36}
		& Fast AT~\cite{fast-free}
		& 0.0
		& 13.9G
		& 76H
		& 72.51 & 53.12 & 52.76
		& \textbf{71.20} & 33.02 & 32.84  \\ 
		&
		& 
		& Random
		& 0.4
		& 7.8G
		& \textbf{51H}
		& 70.46 & 51.27 & 51.03
		& 66.62 & 28.95 & 28.75  \\ 
		& 
		&
		& AGAT
		& 0.03
		& \textbf{7.7G}
		& 53H
		& \textbf{72.69} & \textbf{53.66} & \textbf{53.48}
		& 71.06 & \textbf{33.46} & \textbf{33.17}  \\ 
		\midrule
		\multirow{3}{*}{LeViT-128~\cite{levit}} 
		& \multirow{3}{*}{7.8M} 
		& \multirow{3}{*}{9}
		& Fast AT~\cite{fast-free}
		& 0.0
		& 0.4G
		& 12H
		& \textbf{67.30} & 45.00 & 44.87
		& 64.66 & \textbf{32.11} & \textbf{32.09}  \\ 
		& 
		&
		& Random
		& 0.4
		& \textbf{0.2G}
		& \textbf{10H}
		& 58.14 & 30.62 & 30.52
		& 54.94 & 28.05 & 28.01  \\ 
		& 
		&
		& AGAT
		& 0.15
		& \textbf{0.2G}
		& \textbf{10H}
		& 67.19 & \textbf{45.30} & \textbf{45.21}
		& \textbf{64.98} & 32.02 & 31.88  \\ 
		\midrule
		\multirow{3}{*}{LeViT-256~\cite{levit}} 
		& \multirow{3}{*}{11.0M} 
		& \multirow{3}{*}{12}
		& Fast AT~\cite{fast-free}
		& 0.0
		& 0.6G
		& 15H
		& 68.69 & 46.94 & 46.89
		& \textbf{66.24} & \textbf{33.81} & \textbf{33.62}  \\ 
		& 
		&
		& Random
		& 0.4
		& \textbf{0.4G}
		& \textbf{12H}
		& 60.71 & 31.45 & 31.18
		& 57.64 & 29.89 & 29.66  \\ 
		& 
		&
		& AGAT
		& 0.1
		& \textbf{0.4G}
		& 13H
		& \textbf{68.90} & \textbf{47.37} & \textbf{47.06}
		& 65.98 & 33.12 & 33.05  \\ 
		\midrule
		\multirow{3}{*}{LeViT-384~\cite{levit}} 
		& \multirow{3}{*}{39.0M} 
		& \multirow{3}{*}{12}
		& Fast AT~\cite{fast-free}
		& 0.0
		& 2.35G
		& 28H
		& \textbf{70.01} & 46.24 & 46.13
		& 65.38 & 31.22 & 31.04  \\ 
		& 
		&
		& Random
		& 0.4
		& \textbf{1.3G}
		& \textbf{20H}
		& 63.70 & 33.74 & 33.26
		& 60.11 & 28.51 & 28.02  \\ 
		& 
		&
		& AGAT
		& 0.1
		& \textbf{1.3G}
		& 22H
		& 69.73 & \textbf{48.80} & \textbf{48.59}
		& \textbf{67.02} & \textbf{33.48} & \textbf{33.35}  \\ 
		\midrule
		\multirow{3}{*}{ViT-Tiny~\cite{vit}} 
		& \multirow{3}{*}{5.1M} 
		& \multirow{3}{*}{12}
		& Fast AT~\cite{fast-free}
		& 0.0
		& 1.1G
		& 20H
		& 69.09 & 48.32 & 48.28 
		& \textbf{64.03} & \textbf{31.68} & \textbf{31.20}  \\ 
		& 
		&
		& Random
		& 0.4
		& \textbf{0.6G}
		& \textbf{13H}
		& 61.11 & 31.58 & 31.16
		& 58.10 & 27.43 & 27.25  \\ 
		& 
		&
		& AGAT
		& 0.1
		& \textbf{0.6G}
		& \textbf{13H}
		& \textbf{69.64} & \textbf{48.50} & \textbf{48.46}
		& 63.02 & 31.17 & 31.04 \\ 
        \midrule
        \multirow{3}{*}{ViT-Small~\cite{vit}} 
        & \multirow{3}{*}{22M} 
        & \multirow{3}{*}{12}
		& Fast AT~\cite{fast-free}
		& 0.0
		& 4.6G
		& 47H
		& \textbf{71.37} & \textbf{49.49} & \textbf{49.41}
		& \textbf{66.92} & \textbf{34.07} & \textbf{33.82}  \\ 
		& 
		&
		& Random
		& 0.4
		& 2.7G
		& 33H
		& 63.62 & 33.58 & 33.29
		& 60.18 & 29.10 & 28.91  \\ 
		& 
		&
		& AGAT
		& 0.1
		& \textbf{2.6G}
		& \textbf{32H}
		& 70.62 & 49.00 & 48.85
		& 66.10 & 33.62 & 33.40  \\ 
        \midrule
		\multirow{3}{*}{ViT-Base~\cite{vit}} 
		& \multirow{3}{*}{86M} 
		& \multirow{3}{*}{12}
		& Fast AT~\cite{fast-free}
		& 0.0
		& 17.5G
		& 86H
		& 70.31 & 50.55 & 50.06
		& 65.18 & 33.59 & 33.39  \\ 
		& 
		&
		& Random
		& 0.4
		& 10.5G
		& \textbf{55H}
		& 65.80 & 37.04 & 36.81
		& 61.16 & 30.07 & 30.01  \\ 
		& 
		&
		& AGAT
		& 0.1
		& \textbf{10.1G}
		& 56H
		& \textbf{70.41} & \textbf{51.23} & \textbf{51.11}
		& \textbf{67.93} & \textbf{34.94} & \textbf{34.78}  \\ 
		\bottomrule
	\end{tabular}}

\end{table*}

\noindent\textbf{Vision Transformers} Our AGAT can be directly used on the self-attention model and most of its variants. We apply our AGAT to three commonly-used models ViT~\cite{vit}, CaiT~\cite{cait}, and LeViT~\cite{levit}. All the three models are built on the original self-attention module and can fully reveal the effectiveness of our AGAT. In future work, we will explore combining our AGAT with more sophisticated attention mechanism like window attention~\cite{swin} or multi-scale attention~\cite{DBLP:journals/corr/crossformer}.

\noindent\textbf{Training Algorithm} 
For vision transformers, we compare our AGAT with FastAT and the random dropping strategy (Random). we also provide results of Free AT~\cite{for-free} and Grad Align~\cite{grad-align} algorithms on the ResNet~\cite{resnetv1} models.

\subsection{Improved Efficiency of Adversarial Training on ImageNet}

We present the performance of AGAT in Table~\ref{table:main}. Because AGAT drops a static rate of embeddings for each self-attention layer, the depth of the ViTs will decide the total number of dropped features. Thus, we adjust the dropping rate for ViTs with different numbers of blocks so that the complexity of the feed-forward process will be approximately reduced by $40\%$ of the baseline. For instance, we set the dropping rate to $0.1$ for models with $12$ blocks but $0.03$ for models with $36$ blocks. For the baseline of randomly dropping input patches, we can always set the dropping rate to $0.4$ since the total amount of saved computation will not be affected by the number of blocks.

As shown by Table~\ref{table:main}, for each vision transformer, the Fast AT baseline achieves high robustness but is time-consuming, while the
Random dropping strategy saves training time but achieves inferior robustness. 
In contrast, AGAT achieves comparable robustness with Fast AT using much less training time. Particularly, on the ViT-Base model, AGAT achieves similar robustness to Fast AT but only takes $65\%$ of training time.
Meanwhile, since slim models such as ViT-Tiny and LeViT-128 do not possess the same level of model redundancy as their large-sized counterparts, the robustness of these slim models degrades more dramatically than larger ones when we randomly drop patches. When trained with AGAT, the robustness of slim models matches the plain Fast AT .

\begin{table*}[t!]
	\caption{Evaluation of Adversarially Trained Models under Various Attacks. Robust accuracy of adversarially trained models is reported in this table. The robust accuracy achieved by our AGAT is comparable to that by Fast-AT under various attack evaluations and different perturbation ranges.}
	\centering
	\scriptsize
	\label{table:attacks}
	\resizebox{\textwidth}{!}{
	\begin{tabular}{ cc p{\s}p{\s}p{\s}p{\s}p{\s} p{\s}p{\s}p{\s}p{\s}p{\s} } 
		\toprule
		\multirow{3}{*}{Method} & \multirow{3}{*}{Model}
		& \multicolumn{5}{c}{$\epsilon=2/255$}
		& \multicolumn{5}{c}{$\epsilon=4/255$}\\
		\cmidrule(r){3-7}
		\cmidrule(r){8-12}
		& 
		& PGD
		& C\&W
		& APGD-CE
		& APGD-DLR
		& Square
		& PGD
		& C\&W
		& APGD-CE
		& APGD-DLR 
		& Square \\
		\midrule
		\multirow{2}{*}{ViT-Tiny~\cite{vit}} 
		& Fast AT
		& 48.28 & 48.02  & 48.20 & 48.05 & 58.21
		& \textbf{31.20} & \textbf{31.09}  & \textbf{31.11} & \textbf{31.00} & \textbf{37.97}\\ 
		& AGAT
		& \textbf{48.46} & \textbf{48.17}  & \textbf{48.37} & \textbf{48.24} & \textbf{58.33}
		& 31.04 & 31.03 & 31.01 & 30.85 & 37.84\\ 
		\midrule
		\multirow{2}{*}{ViT-Base~\cite{vit}} 
		& Fast AT
		& 50.06 & 49.87  & 50.01 & 49.52 & 59.28
		& 33.39 & 33.36  & 33.28 & 33.10 & 40.81\\ 
		& AGAT
		& \textbf{51.11} & \textbf{50.84}  & \textbf{51.23} & \textbf{50.45} & \textbf{61.12}
		& \textbf{34.78} & \textbf{34.42}  & \textbf{34.54} & \textbf{34.20} & \textbf{42.24} \\ 
		\midrule
		\multirow{2}{*}{CaiT-XXS24~\cite{cait}} 
		& Fast AT
		& \textbf{54.26} & \textbf{54.19}  & \textbf{54.20} & \textbf{54.13} & \textbf{63.22}
		& \textbf{32.09} & \textbf{32.01}  & \textbf{32.08} & \textbf{31.90} & \textbf{38.02}\\ 
		& AGAT
		& 54.08 & 53.88  & 54.00 & 53.98  & 62.35
		& 31.39 & 31.22  & 31.20 & 31.11  & 37.48\\ 
		\midrule
		\multirow{2}{*}{CaiT-S36~\cite{cait}} 
		& Fast AT
		& 52.76 & 52.51  & 52.70 & 52.53  & 61.75
		& 32.84 & 32.73  & 32.79 & 32.61  & 39.71\\ 
		& AGAT
		& \textbf{53.48} & \textbf{53.13}  & \textbf{53.37} & \textbf{53.02}  & \textbf{62.86}
		& \textbf{33.17} & \textbf{33.08}  & \textbf{33.10} & \textbf{33.00} & \textbf{40.12}\\ 
		\midrule 
		\multirow{2}{*}{LeViT-128~\cite{levit}} 
		& Fast AT
		& 44.87 & 44.81  & 44.73 & 44.60 & 56.89
		& \textbf{32.09} & \textbf{32.03}  & \textbf{32.02} & \textbf{31.95} & \textbf{38.90} \\ 
		& AGAT
		& \textbf{45.21} & \textbf{45.19}  & \textbf{45.19} & \textbf{45.06} & \textbf{57.12}
		& 31.88 & 31.77  & 31.87 & 31.76  & 38.61\\ 
		\midrule
		\multirow{2}{*}{LeViT-384~\cite{levit}} 
		& Fast AT
		& 46.13 & 45.98  & 46.02 & 45.93  & 56.39
		& 31.25 & 31.00  & 31.22 & 30.82  & 37.99\\ 
		& AGAT
		& \textbf{48.59} & \textbf{48.26}  & \textbf{48.58} & \textbf{48.49}  & \textbf{58.90}
		& \textbf{33.35} & \textbf{33.24}  & \textbf{33.29} & \textbf{33.17} & \textbf{40.22}\\ 
		\bottomrule
	\end{tabular}}
\end{table*}

\subsection{Ablation Study}

\subsubsection{Results under Various Attacks.}

Adversarial robustness is known to be hard to examine. Several defensive algorithms were found to be vulnerable to tailored attacks. One of the most important and typical representatives of such phenomena is the obfuscated gradient problem. Our AGAT does not fall into this category, considering the algorithm neither utilizes any stochastic process nor hampers the gradient computation. In fact, our AGAT only takes effect on the training stage and does not modify any procedure during evaluation. To further show the robustness of the learned ViTs, we present the robust accuracy of our learned models under various different attacks in Table~\ref{table:attacks}. We select the attacking criteria of C\&W~\cite{cw}, Square~\cite{square}, APGD-CE~\cite{autoattack}, and APGD-DLR~\cite{autoattack}. 
The AGAT models achieve the same level of robustness as Fast AT.

\subsubsection{Robustness and Dropping Rate.}

We cross-validate the rate of dropping of our AGAT in Table~\ref{table:rate}. To provide a clear comparison with the random dropping strategy, we compare their learned robust accuracy when both dropping strategies reduce approximately the same amount of computation. Due to the difference between the two algorithms (layer-wise vs input-wise), the actual learning rates are different across the two methods. It can be told that AGAT can maintain the learned robustness in a wide range of dropping rates. In contrast, the random dropping strategy significantly degrades the performance, especially for the slim model of ViT-Tiny. Dropping more than $40\%$ computation will bring obvious degradation on robustness, even for AGAT. Thus, we consider this dropping rate as a good trade-off between effectiveness and efficiency.

\begin{table*}[t!]
	\caption{Ablation Study on Dropping Strategy. We compare our attention-guided dropping strategy with random dropping. Ours outperforms random dropping constantly in different dropping rates.}
	\centering
	\label{table:rate}
	\scriptsize
	\setlength\tabcolsep{0.12cm}
	\begin{tabular}{ cccccccccccc } 
		\toprule
		& &
		& \multicolumn{4}{c}{$\epsilon=2/255$}
		& \multicolumn{4}{c}{$\epsilon=4/255$}\\
		\cmidrule(r){4-7}
		\cmidrule(r){8-11}
		  Model
		& Method
		& Attack
		& $0\%$
		& $20\%$
		& $40\%$
		& $60\%$
		& $0\%$
		& $20\%$
		& $40\%$
		& $60\%$\\
		\midrule
		\multirow{2}{*}{ViT-Tiny~\cite{vit}} 
		& Random
		& PGD
		& 48.28 & 39.02 & 31.16 & 21.01  
		& 31.20 & 29.64 & 27.25 & 24.11 \\ 
		& AGAT
		& PGD
		& 48.28 & \textbf{48.15} & \textbf{48.46} & \textbf{46.60}  
		& 31.20 & \textbf{31.15} & \textbf{31.04} & \textbf{28.13} \\ 
		\midrule
		\multirow{2}{*}{ViT-Small~\cite{vit}} 
		& Random
		& PGD
		& 49.41 & 40.62 & 33.29 & 28.81  
		& 33.82 & 30.13 & 28.91 & 25.25 \\ 
		& AGAT
		& PGD
		& 49.41 & \textbf{49.43} & \textbf{48.85} & \textbf{47.90}  
		& 33.82 & \textbf{33.29} & \textbf{33.40} & \textbf{32.10} \\ 
		\midrule
		\multirow{2}{*}{ViT-Base~\cite{vit}} 
		& Random
		& PGD
		& 49.26 & 45.02 & 36.81 & 33.17  
		& 33.39 & 32.61 & 30.01 & 28.86 \\ 
		& AGAT
		& PGD
		& 49.26 & \textbf{49.80} & \textbf{50.02} & \textbf{48.20}  
		& 33.39 & \textbf{34.15} & \textbf{34.78} & \textbf{32.90} \\ 

		\bottomrule
	\end{tabular}
\end{table*}

\subsubsection{Visualization.}

\begin{figure*}[!t]
\centering
\label{fig:vis}
\includegraphics[width=\linewidth]{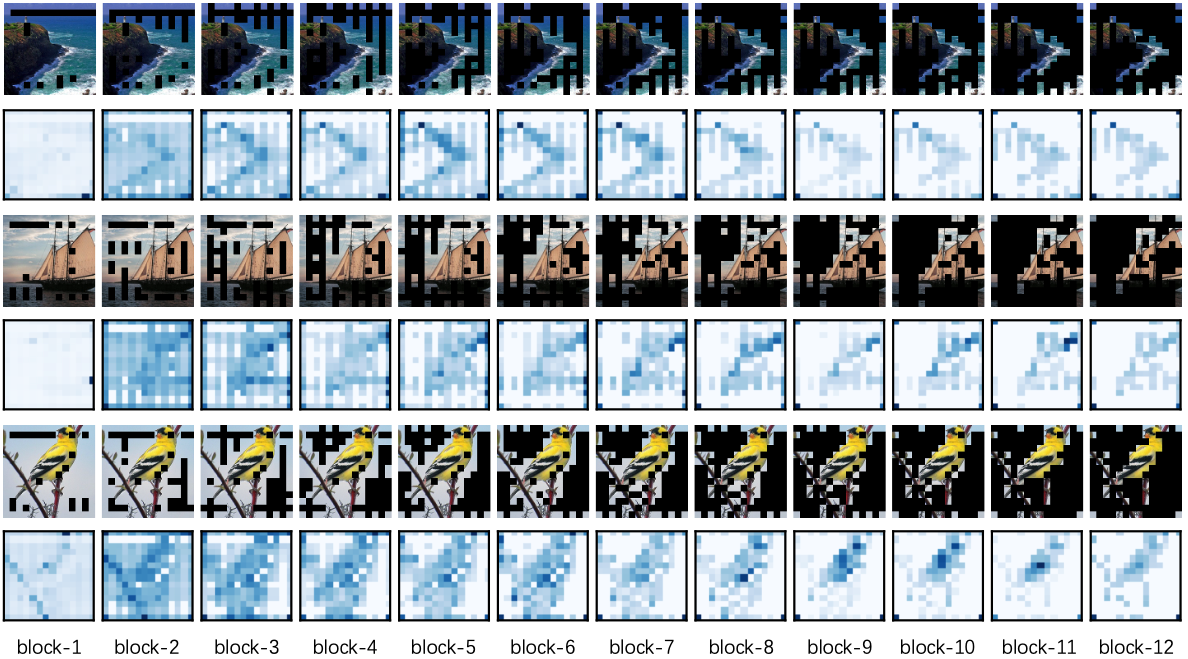}
\caption{Visualizations of dropped image patches and the distribution of attention. AGAT mainly cuts off the computation of non-object embeddings and thus maintains the performance and gradient computation.}
\end{figure*}

To better get an insight of how AGAT takes effect, we visualize the internal results of the ViT-Base model. For each of the 12 blocks in ViT-Base, we show the position of the dropped embeddings by masking out the corresponding image patches. In Fig.~\ref{fig:vis}, the position of the dropped embedding mainly concentrates on the relatively unimportant positions like background, while the patches of the main object are mostly kept. This indicates that our AGAT successfully guards the feed-forward procedure and thus secures the generation of adversarial examples. We also visualize the corresponding value of attention for each patch. The darkness of each patch position indicates how much the corresponding embedding of this patch influences the other embeddings. For each block, we normalize all the values of attention by dividing the maximum value of attention. This visualization also demonstrates that the dropping rate of AGAT gets to cover the embeddings on the position of the main object. Therefore, dropping rates larger than the chosen value may lose crucial information for inference and thus hamper the generation of adversarial examples for training.

\section{Conclusions}

Adversarial training is one of the most effective defense methods to boost the adversarial robustness of models. However, it is computationally expensive, even after many efforts have been made to address it. The emergence of ViTs, whose computational cost increases quadratically with the number of input patches, makes adversarial training more challenging.
In this work, we first thoroughly examined the most popular fast adversarial training on various ViTs. Our investigation shows that ViT achieves higher robust accuracy than ResNet, while it does suffer from a large computation burden, as expected. 
Our further exploration showed that random input patch dropping can accelerate and stabilize the adversarial training, which, however, sacrifices the final robust accuracy. 
To overcome the dilemma, we proposed an Attention-Guided Adversarial Training (AGAT) mechanism based on the specialty of the self-attention mechanism. Our AGAT leverages the attention to guide the patch dropping process, which accelerates the adversarial training significantly and maintains the high robust accuracy of ViTs. We hope that this work can serve the community as a baseline for research on efficient adversarial training on vision transformers.

~\\

\noindent\textbf{Acknowledgements.} This work was supported in part by The National Key Research and Development Program of China (Grant Nos: 2018AAA0101400), in part by The National Nature Science Foundation of China (Grant Nos: 62036009, U1909203, 61936006, 62133013), in part by Innovation Capability Support Program of Shaanxi (Program No. 2021TD-05).

\clearpage

\bibliographystyle{splncs04}
\bibliography{references}

\begin{thebibliography}{10}
\providecommand{\url}[1]{\texttt{#1}}
\providecommand{\urlprefix}{URL }
\providecommand{\doi}[1]{https://doi.org/#1}

\bibitem{aldahdooh2021reveal}
Aldahdooh, A., Hamidouche, W., Deforges, O.: Reveal of vision transformers
  robustness against adversarial attacks. arXiv:2106.03734  (2021)

\bibitem{square}
Andriushchenko, M., Croce, F., Flammarion, N., Hein, M.: Square attack: {A}
  query-efficient black-box adversarial attack via random search. In: Vedaldi,
  A., Bischof, H., Brox, T., Frahm, J. (eds.) ECCV. Lecture Notes in Computer
  Science, vol. 12368, pp. 484--501. Springer (2020).
  \doi{10.1007/978-3-030-58592-1\_29},
  \url{https://doi.org/10.1007/978-3-030-58592-1\_29}

\bibitem{andriushchenko2020understanding}
Andriushchenko, M., Flammarion, N.: Understanding and improving fast
  adversarial training. NeurIPS  (2020)

\bibitem{grad-align}
Andriushchenko, M., Flammarion, N.: Understanding and improving fast
  adversarial training. In: Larochelle, H., Ranzato, M., Hadsell, R., Balcan,
  M., Lin, H. (eds.) Advances in Neural Information Processing Systems 33:
  Annual Conference on Neural Information Processing Systems 2020, NeurIPS
  2020, December 6-12, 2020, virtual (2020),
  \url{https://proceedings.neurips.cc/paper/2020/hash/b8ce47761ed7b3b6f48b583350b7f9e4-Abstract.html}

\bibitem{obfuscated}
Athalye, A., Carlini, N., Wagner, D.A.: Obfuscated gradients give a false sense
  of security: Circumventing defenses to adversarial examples. In: ICML.
  Proceedings of Machine Learning Research, vol.~80, pp. 274--283. {PMLR}
  (2018)

\bibitem{DBLP:journals/corr/abs-2204-00993}
Bai, J., Yuan, L., Xia, S., Yan, S., Li, Z., Liu, W.: Improving vision
  transformers by revisiting high-frequency components. CoRR
  \textbf{abs/2204.00993} (2022). \doi{10.48550/arXiv.2204.00993},
  \url{https://doi.org/10.48550/arXiv.2204.00993}

\bibitem{bai2021transformers}
Bai, Y., Mei, J., Yuille, A., Xie, C.: Are transformers more robust than cnns?
  arXiv:2111.05464  (2021)

\bibitem{bai2021are}
Bai, Y., Mei, J., Yuille, A., Xie, C.: Are transformers more robust than
  {CNN}s? In: Beygelzimer, A., Dauphin, Y., Liang, P., Vaughan, J.W. (eds.)
  Advances in Neural Information Processing Systems (2021),
  \url{https://openreview.net/forum?id=hbHkvGBZB9}

\bibitem{vit-rob-5}
Benz, P., Ham, S., Zhang, C., Karjauv, A., Kweon, I.S.: Adversarial robustness
  comparison of vision transformer and mlp-mixer to cnns. CoRR
  \textbf{abs/2110.02797} (2021), \url{https://arxiv.org/abs/2110.02797}

\bibitem{benz2021adversarial}
Benz, P., Ham, S., Zhang, C., Karjauv, A., Kweon, I.S.: Adversarial robustness
  comparison of vision transformer and mlp-mixer to cnns. arXiv preprint
  arXiv:2110.02797  (2021)

\bibitem{vit-rob-1}
Bhojanapalli, S., Chakrabarti, A., Glasner, D., Li, D., Unterthiner, T., Veit,
  A.: Understanding robustness of transformers for image classification. CoRR
  \textbf{abs/2103.14586} (2021), \url{https://arxiv.org/abs/2103.14586}

\bibitem{bhojanapalli2021understanding}
Bhojanapalli, S., Chakrabarti, A., Glasner, D., Li, D., Unterthiner, T., Veit,
  A.: Understanding robustness of transformers for image classification.
  arXiv:2103.14586  (2021)

\bibitem{DBLP:conf/nips/BrownMRSKDNSSAA20}
Brown, T.B., Mann, B., Ryder, N., Subbiah, M., Kaplan, J., Dhariwal, P.,
  Neelakantan, A., Shyam, P., Sastry, G., Askell, A., Agarwal, S.,
  Herbert{-}Voss, A., Krueger, G., Henighan, T., Child, R., Ramesh, A.,
  Ziegler, D.M., Wu, J., Winter, C., Hesse, C., Chen, M., Sigler, E., Litwin,
  M., Gray, S., Chess, B., Clark, J., Berner, C., McCandlish, S., Radford, A.,
  Sutskever, I., Amodei, D.: Language models are few-shot learners. In: Neural
  Information Processing Systems, {NeurIPS} (2020)

\bibitem{cw}
Carlini, N., Wagner, D.A.: Towards evaluating the robustness of neural
  networks. In: SP. pp. 39--57. {IEEE} Computer Society (2017)

\bibitem{CaronTMJMBJ21}
Caron, M., Touvron, H., Misra, I., J{\'{e}}gou, H., Mairal, J., Bojanowski, P.,
  Joulin, A.: Emerging properties in self-supervised vision transformers. In:
  2021 {IEEE/CVF} International Conference on Computer Vision, {ICCV} 2021,
  Montreal, QC, Canada, October 10-17, 2021. pp. 9630--9640. {IEEE} (2021).
  \doi{10.1109/ICCV48922.2021.00951},
  \url{https://doi.org/10.1109/ICCV48922.2021.00951}

\bibitem{drop-patch-3}
Chen, T., Cheng, Y., Gan, Z., Yuan, L., Zhang, L., Wang, Z.: Chasing sparsity
  in vision transformers: An end-to-end exploration. CoRR
  \textbf{abs/2106.04533} (2021), \url{https://arxiv.org/abs/2106.04533}

\bibitem{DBLP:journals/corr/abs-2104-13840}
Chu, X., Tian, Z., Wang, Y., Zhang, B., Ren, H., Wei, X., Xia, H., Shen, C.:
  Twins: Revisiting spatial attention design in vision transformers. CoRR
  \textbf{abs/2104.13840} (2021)

\bibitem{autoattack}
Croce, F., Hein, M.: Reliable evaluation of adversarial robustness with an
  ensemble of diverse parameter-free attacks. In: ICML (2020)

\bibitem{DBLP:conf/naacl/DevlinCLT19}
Devlin, J., Chang, M., Lee, K., Toutanova, K.: {BERT:} pre-training of deep
  bidirectional transformers for language understanding. In: Conference of the
  North American Chapter of the Association for Computational Linguistics,
  {NAACL}. pp. 4171--4186 (2019)

\bibitem{dhillon2018stochastic}
Dhillon, G.S., Azizzadenesheli, K., Lipton, Z.C., Bernstein, J., Kossaifi, J.,
  Khanna, A., Anandkumar, A.: Stochastic activation pruning for robust
  adversarial defense. ICLR  (2018)

\bibitem{vit}
Dosovitskiy, A., Beyer, L., Kolesnikov, A., Weissenborn, D., Zhai, X.,
  Unterthiner, T., Dehghani, M., Minderer, M., Heigold, G., Gelly, S.,
  Uszkoreit, J., Houlsby, N.: An image is worth 16x16 words: Transformers for
  image recognition at scale. In: International Conference on Learning
  Representations, {ICLR} (2021)

\bibitem{fu2021patch}
Fu, Y., Zhang, S., Wu, S., Wan, C., Lin, Y.: Patch-fool: Are vision
  transformers always robust against adversarial perturbations? In:
  International Conference on Learning Representations (2022)

\bibitem{fgsm}
Goodfellow, I.J., Shlens, J., Szegedy, C.: Explaining and harnessing
  adversarial examples. In: Bengio, Y., LeCun, Y. (eds.) ICLR (2015),
  \url{http://arxiv.org/abs/1412.6572}

\bibitem{levit}
Graham, B., El{-}Nouby, A., Touvron, H., Stock, P., Joulin, A., J{\'{e}}gou,
  H., Douze, M.: Levit: a vision transformer in convnet's clothing for faster
  inference. CoRR  \textbf{abs/2104.01136} (2021),
  \url{https://arxiv.org/abs/2104.01136}

\bibitem{vit-rob-4}
Gu, J., Tresp, V., Qin, Y.: Are vision transformers robust to patch
  perturbations? CoRR  \textbf{abs/2111.10659} (2021),
  \url{https://arxiv.org/abs/2111.10659}

\bibitem{gu2021vision}
Gu, J., Tresp, V., Qin, Y.: Are vision transformers robust to patch
  perturbations? In: arXiv preprint arXiv:2111.10659 (2021)

\bibitem{guo2017countering}
Guo, C., Rana, M., Cisse, M., Van Der~Maaten, L.: Countering adversarial images
  using input transformations. ICLR  (2018)

\bibitem{resnetv1}
He, K., Zhang, X., Ren, S., Sun, J.: Deep residual learning for image
  recognition. In: CVPR. pp. 770--778 (2016). \doi{10.1109/CVPR.2016.90},
  \url{https://doi.org/10.1109/CVPR.2016.90}

\bibitem{hu2021inheritance}
Hu, H., Lu, X., Zhang, X., Zhang, T., Sun, G.: Inheritance attention
  matrix-based universal adversarial perturbations on vision transformers. IEEE
  Signal Processing Letters  \textbf{28},  1923--1927 (2021)

\bibitem{Jia_2022_CVPR}
Jia, X., Zhang, Y., Wu, B., Ma, K., Wang, J., Cao, X.: Las-at: Adversarial
  training with learnable attack strategy. In: Proceedings of the IEEE/CVF
  Conference on Computer Vision and Pattern Recognition (CVPR). pp.
  13398--13408 (June 2022)

\bibitem{jia2022boosting}
Jia, X., Zhang, Y., Wu, B., Wang, J., Cao, X.: Boosting fast adversarial
  training with learnable adversarial initialization. IEEE Transactions on
  Image Processing  (2022)

\bibitem{joshi2021adversarial}
Joshi, A., Jagatap, G., Hegde, C.: Adversarial token attacks on vision
  transformers. arXiv:2110.04337  (2021)

\bibitem{ckpt}
Kim, H., Lee, W., Lee, J.: Understanding catastrophic overfitting in
  single-step adversarial training. In: Thirty-Fifth {AAAI} Conference on
  Artificial Intelligence, {AAAI} 2021, Thirty-Third Conference on Innovative
  Applications of Artificial Intelligence, {IAAI} 2021, The Eleventh Symposium
  on Educational Advances in Artificial Intelligence, {EAAI} 2021, Virtual
  Event, February 2-9, 2021. pp. 8119--8127. {AAAI} Press (2021),
  \url{https://ojs.aaai.org/index.php/AAAI/article/view/16989}

\bibitem{fat-over-2}
Kim, H., Lee, W., Lee, J.: Understanding catastrophic overfitting in
  single-step adversarial training. In: Thirty-Fifth {AAAI} Conference on
  Artificial Intelligence, {AAAI} 2021, Thirty-Third Conference on Innovative
  Applications of Artificial Intelligence, {IAAI} 2021, The Eleventh Symposium
  on Educational Advances in Artificial Intelligence, {EAAI} 2021, Virtual
  Event, February 2-9, 2021. pp. 8119--8127. {AAAI} Press (2021),
  \url{https://ojs.aaai.org/index.php/AAAI/article/view/16989}

\bibitem{adamw}
Kingma, D.P., Ba, J.: Adam: {A} method for stochastic optimization. In: Bengio,
  Y., LeCun, Y. (eds.) 3rd International Conference on Learning
  Representations, {ICLR} 2015, San Diego, CA, USA, May 7-9, 2015, Conference
  Track Proceedings (2015), \url{http://arxiv.org/abs/1412.6980}

\bibitem{cifar}
Krizhevsky, A., Hinton, G., et~al.: Learning multiple layers of features from
  tiny images  (2009)

\bibitem{mnist}
LeCun, Y., Cortes, C.: {MNIST} handwritten digit database  (2010),
  \url{http://yann.lecun.com/exdb/mnist/}

\bibitem{liang2020efficient}
Liang, S., Wei, X., Yao, S., Cao, X.: Efficient adversarial attacks for visual
  object tracking. In: European Conference on Computer Vision. pp. 34--50.
  Springer (2020)

\bibitem{liang2021parallel}
Liang, S., Wu, B., Fan, Y., Wei, X., Cao, X.: Parallel rectangle flip attack: A
  query-based black-box attack against object detection. In: Proceedings of the
  IEEE/CVF International Conference on Computer Vision. pp. 7697--7707 (2021)

\bibitem{liao2018defense}
Liao, F., Liang, M., Dong, Y., Pang, T., Hu, X., Zhu, J.: Defense against
  adversarial attacks using high-level representation guided denoiser. In:
  CVPR. pp. 1778--1787 (2018)

\bibitem{LiuJLC11}
Liu, W., Jiang, Y., Luo, J., Chang, S.: Noise resistant graph ranking for
  improved web image search. In: The 24th {IEEE} Conference on Computer Vision
  and Pattern Recognition, {CVPR} 2011, Colorado Springs, CO, USA, 20-25 June
  2011. pp. 849--856. {IEEE} Computer Society (2011).
  \doi{10.1109/CVPR.2011.5995315},
  \url{https://doi.org/10.1109/CVPR.2011.5995315}

\bibitem{swin}
Liu, Z., Lin, Y., Cao, Y., Hu, H., Wei, Y., Zhang, Z., Lin, S., Guo, B.: Swin
  transformer: Hierarchical vision transformer using shifted windows. CoRR
  \textbf{abs/2103.14030} (2021)

\bibitem{pgd}
Madry, A., Makelov, A., Schmidt, L., Tsipras, D., Vladu, A.: Towards deep
  learning models resistant to adversarial attacks. In: ICLR. OpenReview.net
  (2018), \url{https://openreview.net/forum?id=rJzIBfZAb}

\bibitem{mahmood2021robustness}
Mahmood, K., Mahmood, R., Van~Dijk, M.: On the robustness of vision
  transformers to adversarial examples. In: Proceedings of the IEEE/CVF
  International Conference on Computer Vision. pp. 7838--7847 (2021)

\bibitem{mao2021towards}
Mao, X., Qi, G., Chen, Y., Li, X., Duan, R., Ye, S., He, Y., Xue, H.: Towards
  robust vision transformer. arXiv:2105.07926  (2021)

\bibitem{mao2021rethinking}
Mao, X., Qi, G., Chen, Y., Li, X., Ye, S., He, Y., Xue, H.: Rethinking the
  design principles of robust vision transformer. arXiv:2105.07926  (2021)

\bibitem{meng2017magnet}
Meng, D., Chen, H.: Magnet: a two-pronged defense against adversarial examples.
  In: Proceedings of the 2017 ACM SIGSAC Conference on Computer and
  Communications Security. pp. 135--147 (2017)

\bibitem{mu2021defending}
Mu, N., Wagner, D.: Defending against adversarial patches with robust
  self-attention. In: ICML 2021 Workshop on Uncertainty and Robustness in Deep
  Learning (2021)

\bibitem{naseer2021intriguing}
Naseer, M.M., Ranasinghe, K., Khan, S.H., Hayat, M., Shahbaz~Khan, F., Yang,
  M.H.: Intriguing properties of vision transformers. Advances in Neural
  Information Processing Systems  \textbf{34} (2021)

\bibitem{drop-patch-4}
Pan, B., Jiang, Y., Panda, R., Wang, Z., Feris, R., Oliva, A.:
  Ia-red\({}^{\mbox{2}}\): Interpretability-aware redundancy reduction for
  vision transformers. CoRR  \textbf{abs/2106.12620} (2021),
  \url{https://arxiv.org/abs/2106.12620}

\bibitem{park2021reliably}
Park, G.Y., Lee, S.W.: Reliably fast adversarial training via latent
  adversarial perturbation. ICCV  (2021)

\bibitem{vit-rob-2}
Paul, S., Chen, P.: Vision transformers are robust learners. CoRR
  \textbf{abs/2105.07581} (2021), \url{https://arxiv.org/abs/2105.07581}

\bibitem{paul2021vision}
Paul, S., Chen, P.Y.: Vision transformers are robust learners. arXiv:2105.07581
   (2021)

\bibitem{llr}
Qin, C., Martens, J., Gowal, S., Krishnan, D., Dvijotham, K., Fawzi, A., De,
  S., Stanforth, R., Kohli, P.: Adversarial robustness through local
  linearization. In: NeurIPS (2019)

\bibitem{drop-patch-2}
Rao, Y., Zhao, W., Liu, B., Lu, J., Zhou, J., Hsieh, C.: Dynamicvit: Efficient
  vision transformers with dynamic token sparsification. CoRR
  \textbf{abs/2106.02034} (2021), \url{https://arxiv.org/abs/2106.02034}

\bibitem{over-fitting}
Rice, L., Wong, E., Kolter, J.Z.: Overfitting in adversarially robust deep
  learning. CoRR  \textbf{abs/2002.11569} (2020),
  \url{https://arxiv.org/abs/2002.11569}

\bibitem{imagenet}
Russakovsky, O., Deng, J., Su, H., Krause, J., Satheesh, S., Ma, S., Huang, Z.,
  Karpathy, A., Khosla, A., Bernstein, M., Berg, A.C., Fei-Fei, L.: Imagenet
  large scale visual recognition challenge (2015)

\bibitem{DBLP:journals/ijcv/RussakovskyDSKS15}
Russakovsky, O., Deng, J., Su, H., Krause, J., Satheesh, S., Ma, S., Huang, Z.,
  Karpathy, A., Khosla, A., Bernstein, M.S., Berg, A.C., Li, F.: Imagenet large
  scale visual recognition challenge. International Journal of Computer Vision,
  {IJCV}  (2015)

\bibitem{fat-over-1}
S., V.B., Babu, R.V.: Single-step adversarial training with dropout scheduling.
  In: 2020 {IEEE/CVF} Conference on Computer Vision and Pattern Recognition,
  {CVPR} 2020, Seattle, WA, USA, June 13-19, 2020. pp. 947--956. Computer
  Vision Foundation / {IEEE} (2020). \doi{10.1109/CVPR42600.2020.00103},
  \url{https://openaccess.thecvf.com/content\_CVPR\_2020/html/B.S.\_Single-Step\_Adversarial\_Training\_With\_Dropout\_Scheduling\_CVPR\_2020\_paper.html}

\bibitem{salman2021certified}
Salman, H., Jain, S., Wong, E., Madry, A.: Certified patch robustness via
  smoothed vision transformers. arXiv:2110.07719  (2021)

\bibitem{samangouei2018defense}
Samangouei, P., Kabkab, M., Chellappa, R.: Defense-gan: Protecting classifiers
  against adversarial attacks using generative models. ICLR  (2018)

\bibitem{for-free}
Shafahi, A., Najibi, M., Ghiasi, A., Xu, Z., Dickerson, J.P., Studer, C.,
  Davis, L.S., Taylor, G., Goldstein, T.: Adversarial training for free! In:
  NeurIPS (2019)

\bibitem{shao2021adversarial}
Shao, R., Shi, Z., Yi, J., Chen, P.Y., Hsieh, C.J.: On the adversarial
  robustness of visual transformers. arXiv:2103.15670  (2021)

\bibitem{shi2021decision}
Shi, Y., Han, Y.: Decision-based black-box attack against vision transformers
  via patch-wise adversarial removal. arXiv preprint arXiv:2112.03492  (2021)

\bibitem{song2017pixeldefend}
Song, Y., Kim, T., Nowozin, S., Ermon, S., Kushman, N.: Pixeldefend: Leveraging
  generative models to understand and defend against adversarial examples. ICLR
   (2018)

\bibitem{sriramanan2021towards}
Sriramanan, G., Addepalli, S., Baburaj, A., et~al.: Towards efficient and
  effective adversarial training. NeurIPS  (2021)

\bibitem{intrigue}
Szegedy, C., Zaremba, W., Sutskever, I., Bruna, J., Erhan, D., Goodfellow,
  I.J., Fergus, R.: Intriguing properties of neural networks. In: Bengio, Y.,
  LeCun, Y. (eds.) ICLR (2014), \url{http://arxiv.org/abs/1312.6199}

\bibitem{tang2021robustart}
Tang, S., Gong, R., Wang, Y., Liu, A., Wang, J., Chen, X., Yu, F., Liu, X.,
  Song, D., Yuille, A., et~al.: Robustart: Benchmarking robustness on
  architecture design and training techniques. arXiv preprint arXiv:2109.05211
  (2021)

\bibitem{robustart}
Tang, S., Gong, R., Wang, Y., Liu, A., Wang, J., Chen, X., Yu, F., Liu, X.,
  Song, D., Yuille, A.L., Torr, P.H.S., Tao, D.: Robustart: Benchmarking
  robustness on architecture design and training techniques. CoRR
  \textbf{abs/2109.05211} (2021), \url{https://arxiv.org/abs/2109.05211}

\bibitem{drop-patch-1}
Tang, Y., Han, K., Wang, Y., Xu, C., Guo, J., Xu, C., Tao, D.: Patch slimming
  for efficient vision transformers. CoRR  \textbf{abs/2106.02852} (2021),
  \url{https://arxiv.org/abs/2106.02852}

\bibitem{DBLP:conf/icml/TouvronCDMSJ21}
Touvron, H., Cord, M., Douze, M., Massa, F., Sablayrolles, A., J{\'{e}}gou, H.:
  Training data-efficient image transformers {\&} distillation through
  attention. In: International Conference on Machine Learning, {ICML}.
  vol.~139, pp. 10347--10357 (2021)

\bibitem{deit}
Touvron, H., Cord, M., Douze, M., Massa, F., Sablayrolles, A., J{\'{e}}gou, H.:
  Training data-efficient image transformers {\&} distillation through
  attention. In: Meila, M., Zhang, T. (eds.) Proceedings of the 38th
  International Conference on Machine Learning, {ICML} 2021, 18-24 July 2021,
  Virtual Event. Proceedings of Machine Learning Research, vol.~139, pp.
  10347--10357. {PMLR} (2021),
  \url{http://proceedings.mlr.press/v139/touvron21a.html}

\bibitem{cait}
Touvron, H., Cord, M., Sablayrolles, A., Synnaeve, G., J{\'{e}}gou, H.: Going
  deeper with image transformers. CoRR  \textbf{abs/2103.17239} (2021),
  \url{https://arxiv.org/abs/2103.17239}

\bibitem{DBLP:conf/nips/VaswaniSPUJGKP17}
Vaswani, A., Shazeer, N., Parmar, N., Uszkoreit, J., Jones, L., Gomez, A.N.,
  Kaiser, L., Polosukhin, I.: Attention is all you need. In: Neural Information
  Processing Systems, {NeurIPS}. pp. 5998--6008 (2017)

\bibitem{gat}
Velickovic, P., Cucurull, G., Casanova, A., Romero, A., Li{\`{o}}, P., Bengio,
  Y.: Graph attention networks. In: 6th International Conference on Learning
  Representations, {ICLR} 2018, Vancouver, BC, Canada, April 30 - May 3, 2018,
  Conference Track Proceedings. OpenReview.net (2018),
  \url{https://openreview.net/forum?id=rJXMpikCZ}

\bibitem{vivek2020single}
Vivek, B., Babu, R.V.: Single-step adversarial training with dropout
  scheduling. In: CVPR (2020)

\bibitem{DBLP:journals/corr/abs-2102-12122}
Wang, W., Xie, E., Li, X., Fan, D., Song, K., Liang, D., Lu, T., Luo, P., Shao,
  L.: Pyramid vision transformer: {A} versatile backbone for dense prediction
  without convolutions. CoRR  \textbf{abs/2102.12122} (2021)

\bibitem{DBLP:journals/corr/crossformer}
Wang, W., Yao, L., Chen, L., Lin, B., Cai, D., He, X., Liu, W.: Crossformer: A
  versatile vision transformer hinging on cross-scale attention. In:
  International Conference on Learning Representations (2022),
  \url{https://openreview.net/forum?id=_PHymLIxuI}

\bibitem{WangJZYSL22}
Wang, Z., Jiang, W., Zhu, Y., Yuan, L., Song, Y., Liu, W.: Dynamixer: {A}
  vision {MLP} architecture with dynamic mixing. In: Chaudhuri, K., Jegelka,
  S., Song, L., Szepesv{\'{a}}ri, C., Niu, G., Sabato, S. (eds.) International
  Conference on Machine Learning, {ICML} 2022, 17-23 July 2022, Baltimore,
  Maryland, {USA}. Proceedings of Machine Learning Research, vol.~162, pp.
  22691--22701. {PMLR} (2022),
  \url{https://proceedings.mlr.press/v162/wang22i.html}

\bibitem{wei2019transferable}
Wei, X., Liang, S., Chen, N., Cao, X.: Transferable adversarial attacks for
  image and video object detection. In: Proceedings of the 28th International
  Joint Conference on Artificial Intelligence. pp. 954--960 (2019)

\bibitem{fast-free}
Wong, E., Rice, L., Kolter, J.Z.: Fast is better than free: Revisiting
  adversarial training. CoRR  \textbf{abs/2001.03994} (2020),
  \url{https://arxiv.org/abs/2001.03994}

\bibitem{xie2017mitigating}
Xie, C., Wang, J., Zhang, Z., Ren, Z., Yuille, A.: Mitigating adversarial
  effects through randomization. ICLR  (2018)

\bibitem{drop-patch-5}
Xu, Y., Zhang, Z., Zhang, M., Sheng, K., Li, K., Dong, W., Zhang, L., Xu, C.,
  Sun, X.: Evo-vit: Slow-fast token evolution for dynamic vision transformer.
  CoRR  \textbf{abs/2108.01390} (2021), \url{https://arxiv.org/abs/2108.01390}

\bibitem{yu2021mia}
Yu, Z., Fu, Y., Li, S., Li, C., Lin, Y.: Mia-former: Efficient and robust
  vision transformers via multi-grained input-adaptation. arXiv preprint
  arXiv:2112.11542  (2021)

\bibitem{DBLP:journals/corr/abs-2105-13677}
Zhang, Q., Yang, Y.: Rest: An efficient transformer for visual recognition.
  CoRR  \textbf{abs/2105.13677} (2021)

\end{thebibliography}
\end{document}